\documentclass[a4paper,12pt]{article}
\usepackage{amsmath,amssymb} %数学公式 数学字体 模板
\usepackage{amsfonts} % 模板
\usepackage{graphicx} % 模板
\usepackage{cases}
\usepackage{mathtools}
\usepackage{ragged2e}
\usepackage{array} % 模板
\usepackage{textcomp} % 模板
\usepackage{url} % 模板
\usepackage{verbatim} % 模板
\usepackage[ruled,lined]{algorithm2e}
\usepackage{float}
\usepackage{amsthm}
\usepackage{mathrsfs}
\usepackage[caption=false,font=normalsize,labelfont=sf,textfont=sf]{subfig} % 模板
\usepackage[backend=biber, style=numeric, maxbibnames=99, minbibnames=1, maxnames=99, minnames=1]{biblatex}%chem-angew
\usepackage{xcolor}
\usepackage{microtype}
\microtypesetup{protrusion=true, expansion=true}
\usepackage{footnote}

\begin{document}
\title{Graph Similarity Regularized Softmax for Semi-Supervised Node Classification}
% \author{Yiming Yang, Jun Liu, Wei Wan,~\IEEEmembership{Staff,~IEEE,}
        % <-this % stops a space
% \author{\IEEEauthorblockN{Yiming Yang\IEEEauthorrefmark{a}, Jun Liu\IEEEauthorrefmark{b}, Wei Wan\IEEEauthorrefmark{a}}

% \IEEEauthorblockA{\IEEEauthorrefmark{a} School of Mathematics and Physics, North China Electric Power University, Beijing, China}
% \IEEEauthorblockA{\IEEEauthorrefmark{b} Laboratory of Mathematics and Complex Systems (Ministry of Education of China) School of Mathematical Sciences, Beijing Normal University Beijing, China}
%\IEEEauthorblockA{\{zhangsan\}@XXX.com, \{lisi, wangwu\}@XXX.edu.cn}
% \thanks{This paper was produced by (the IEEE Publication Technology Group. They are in Piscataway, NJ.)}% <-this % stops a space
\author{
Yiming Yang\thanks{Y. Yang is with the School of Mathematics and Physics, North China Electric Power University, Beijing 102200, China(ymyang@ncepu.edu.cn).}
, Jun Liu\thanks{J. Liu is with the Laboratory of Mathematics and Complex Systems (Ministry of Education of China) School of Mathematical Sciences, Beijing Normal University Beijing, China (e-mail: jliu@bnu.edu.cn).}
, Wei Wan\thanks{W. Wan is with the School of Mathematics and Physics, North China Electric Power University, Beijing 102200, China (e-mail: weiwan@ncepu.edu.cn).}
}

\newcommand{\unnumberedfootnote}[1]{%
  \begingroup
  % 重置脚注计数器和编号格式
  \setcounter{footnote}{0} 
  \renewcommand\thefootnote{}%
  \footnotetext{#1}%
  \addtocounter{footnote}{0}% 确保后续脚注编号不受影响
  \endgroup
}

\unnumberedfootnote{The research was supported in part by the National Natural Science Foundation of China under Grant 12301538, Grant 12371527. (Corresponding author: Wei Wan)}

\date{}
%\thanks{Manuscript received (month) (day), (year); revised (August) (16), (2021).}}
% \markboth{IEEE TRANSACTIONS ON KNOWLEDGE AND DATA ENGINEERING} {Yiming Yang, Jun Liu, Wei Wan, 
%\MakeLowercase{\textit{(et al.)}
% : Graph Similarity Regularized Softmax}
% \markboth{(Journal of \LaTeX\ Class Files,~Vol.~14, No.~8, August~2021)}%
% {Yang \MakeLowercase{\textit{et al.}}: Graph Similarity Regularized Softmax}
\maketitle
\begin{abstract}

Graph Neural Networks (GNNs) are powerful deep learning models designed for graph-structured data, demonstrating effectiveness across a wide range of applications.
%and have proven to be highly effective across a wide range of applications. %, including recommendation systems, bioinformatics, and knowledge graphs.
%The semi-supervised node classification problem is a crucial task in GNNs.
The softmax function is the most commonly used classifier for semi-supervised node classification. However, the softmax function lacks spatial information of the graph structure.
In this paper, we propose a graph similarity regularized softmax for GNNs in semi-supervised node classification. By incorporating non-local total variation (TV) regularization into the softmax activation function, we can more effectively capture the spatial information inherent in graphs. The weights in the non-local gradient and divergence operators are determined based on the graph's adjacency matrix. We apply the proposed method into the architecture of GCN and GraphSAGE, testing them on citation and webpage linking datasets, respectively. Numerical experiments demonstrate its good performance in node classification and generalization capabilities.
These results indicate that the graph similarity regularized softmax is effective on both assortative and disassortative graphs.
%In this work, we propose a variational regularized graph neural network (GNNs) for node classification. By incorporating the non-local total variation (NLTV) regularization into the GNNs framework, we can more effectively leverage the  long-distance spatial information inherent in graphs. The weights in the non-local gradient operator and non-local divergence operators are determined based on the adjacency matrix of the graph. Extensive numerical experiments based on the two variants of GNNs have been conducted to evaluate the proposed model, demonstrating its promising performance and generalization capabilities across several datasets. 
\end{abstract}
\textit{Index Terms - } Graph neural networks, Node classification, Regularization, Non-local total variation.

\section{Introduction}

Graph Neural Networks (GNNs) is a type of neural network specifically designed to operate on graph-structured data. GNNs have overcome the Euclidean data limitation inherent in traditional convolutional Neural Networks (CNNs), facilitating applications in domains like machine translation \cite{42}, recommendation systems \cite{5}, bioinformatics \cite{6}, and interactive simulations \cite{7}. 
Semi-supervised node classification is a crucial task in GNNs which employ node features and edge information within graph data to accurately predict node labels, even when only a limited number of labeled nodes are available. %{\color{blue}Graphs used in node classification tasks can be categorized into assortative and disassortative graph based on assortativity coefficient \cite{56}. % assortativity coefficient同配系数
%i.e., the likelihood that nodes sharing the same label are closely positioned within the graph. 
%Graphs with greater assortativity coefficient $(>0)$ are known as assortative graphs, where nodes with higher degree tend to cluster together, such as citation networks. Conversely, disassortative graph are characterized by lower assortativity coefficient $(<0)$, where nodes with higher degree are more likely to near the nodes with lower degree, such as webpage linking networks \cite{55}.}
Graphs used in node classification tasks can be categorized into assortative and disassortative graph \cite{56, 55} based on node homophily \cite{64}. Graphs with high node homophily are known as assortative graphs, where nodes with the same label tend to cluster together, such as citation networks. Conversely, disassortative graph are characterized by lower node homophily, where nodes with the same label are more likely to be distant from each other, such as webpage linking networks.
%For these distinct data types, we identified two effective GNNs models: Graph Convolutional Network (GCN) for assortativty and GraphSAGE for disassortativty.

%The Graph Convolutional Network (GCN) and the GraphSAGE exemplify the archetypal approaches of transductive and inductive learning within the domain of graph neural networks, respectively. 
%Introduced by Kipf and Welling in 2017 \cite{1}, GCN has become a foundational model in the field of graph neural networks. The core idea behind GCN is to define convolution operations on graphs to effectively aggregate information from a node’s neighbors. In the same year that GCN was proposed, GraphSAGE \cite{3} introduced a novel sampling strategy that allows the model to randomly sample a fixed number of neighboring nodes at each layer, and then aggregate information through aggregation functions, thereby improving the efficiency of GCN. The most discussed aggregation functions are mean aggregation and max pooling, both of which have achieved decent performances in node classification tasks. 
Recently, many variants of GNNs have been proposed and attracted increasing attention. Some of GNNs focus on the aggregation of neighbor information. One notable method is the Graph Convolutional Network (GCN) introduced by Kipf and Welling \cite{1}. The core idea behind GCN is to define convolution operations on graphs to effectively learn representations by aggregating information from a node’s neighbors.
GraphSAGE \cite{3} introduced an innovative sampling strategy that allows the model to randomly sample a fixed number of neighboring nodes at each layer, and then aggregate information through predefined aggregation functions. 
The most discussed aggregation functions are mean aggregation and max pooling, both of which have achieved decent performances in node classification tasks. 
%It's worth mentioning that GraphSAGE extended the task processing objects to large-scale graph data. 
Additionally, common methods for addressing the issue of aggregating neighbor information can be broadly divided into strengthening effective neighbor determination and improving the way aggregating the feature weights. Chen \textit{et al.} \cite{12} proposed the Label-Aware GCN framework, which refines the graph structure by increasing positive ratio of neighbors. Wang \textit{et al.} \cite{11} integrated the high-order motif-structure information into feature aggregation. Liu \textit{et al.} \cite{13}, to complement the original feature data, generated more samples in the local neighborhood via data augmentation. Graph Attention Networks (GAT) \cite{2} leverage the attention mechanism to dynamically allocate weights to neighboring nodes based on their similarity, enabling effective information aggregation from distant neighbors across multiple hops. In \cite{14}, Zhang \textit{et al.} further developed this idea by introducing a model with a mask aggregator, which performs a Hadamard product between the feature vector of each neighbor, supporting both node-level and feature-level attention.
%\textbf{(2)Alleviating oversmoothing:} To mitigate the oversmoothing issue, Li \textit{et al.} \cite{29} co-trains a GCN with a random walk model, which could complement the data information in global graph topology and self-trains a GCN used to overcome GCN's localized nature. 
%In addition, DropEdge \cite{30} is proposed to alleviate oversmoothing by randomly removing a certain number of edges during each training epoch. In \cite{31}, Yang \textit{et al.} introduced propagation-regularization into GCN, which enables nodes to capture information from father nodes to avoid oversmoothing.
%\textbf{(3)New mechanisms:} In terms of integrating novel mechanisms into GCNs, in 2018, inspired by GraphSAGE, GAT introduced the attention mechanism into GCNs \cite{2}. Among them, some researchers have introduced diffusion mechanism \cite{17}, knowledge distillation \cite{15} and manifold learning\cite{18} into GCNs, which are also extremely interesting attempts.
%\textbf{(3)Addressing overfitting:} 
%While the efficacy of GNNs is widely acknowledged, the propensity for overfitting remains a prevalent computational challenge in graph representation learning due to the scarcity of training samples. 

Researchers have made numerous attempts to address the overfitting issues in GNNs.
Tian \textit{et al.} \cite{25} employ stochastic transformation and data perturbations at both the node and edge structure levels to regularize the unlabeled and labeled data. Pei \textit{et al.} \cite{28} measure the saliency of each node from a global perspective, specifically the semantic similarities between each node and the graph representation, and then use the learned saliency distribution to regularize GNNs. Some researchers have explored regularization from the perspective of the loss function. Kejani \textit{et al.} \cite{26} propose a global loss function that integrates supervised and unsupervised information, and mitigates overfitting through manifold regularization applied to unsupervised loss. However, the model with a limitation on high-order neighbors. To fully exploit the structured information, Dornaika \cite{27} further extend this method's feasibility by integrating high-order neighbor feature propagation strategy into each GNN layer. In addition, Fu \textit{et al.} utilize p-Laplacian matrix \cite{32} and hypergraph p-Laplacian \cite{45} to explore manifold structural information, thereby preserving local structural information while completing regularization tasks. 
However, most studies in the field are grounded in empirical experience, our work introduces a novel approach based on the variational method.
%However, the majority of these approaches enhance the model's efficacy by incorporating regularization terms into the loss function or by improving the model's existing mechanisms, without directly making adjustment to the its intrinsic architecture.
%{\color{red}However, in contrast to the majority of these approaches that rely on historical experience, we propose a novel framework grounded in variational principles, specifically tailored for applications within the GNNs domain.}

%Total variational (TV) methods, renowned for their efficacy in minimization problems, have gained prominence in image processing \cite{57, 58, 59, 60}. Recent advancements have witnessed innovative applications in this domain. For instance, the variational approach has been adapted to data clustering on weighted graphs by minimizing a Potts model that incorporates regional and edge force terms, particularly effective for data with complex geometric structures \cite{61}. Building on this, Jia \textit{et al.} \cite{62}integrated graph total variation into CNNs by augmenting the softmax activation function with a spatial regularity term, enhancing CNNs with spatial coherence. Subsequent work \cite{33} addressed the limitations of convolution operators in capturing long-range dependencies, facilitating the elimination of isolated regions in image segmentation while preserving fine details. These regularized networks not only achieve superior segmentation results compared to standard CNNs but also exhibits enhanced robustness to noise. 

Most studies employ a single split of train/validation/test for each dataset to evaluate the accuracy in semi-supervised node classification. However, Oleksandr et al. \cite{37} found that different data splits can lead to significantly different rankings of models. They highlight a major risk: models developed using a single split often perform well only with that specific split, failing to evaluate the model’s true generalization capabilities. To address this limitation, they proposed an assessment strategy based on averaging results over 100 random train/validation/test splits for each dataset and 20 random parameter initializations for each split. This approach provides a more accurate evalution of model generalization performance, avoiding the pitfall of selecting a model that overfits to a single fixed test set.

Total Variation (TV) regularization \cite{63} is one of the most widely used methods in image processing, known for its outstanding performance in image restoration tasks. Recently, Fan et al. \cite{62} proposed a framework that integrates traditional variational regularization method into CNNs for semantic image segmentation. In their work, the softmax activation function is reinterpreted as the minimizer of a variational problem, then spatial TV regularization can be effectively incorporated into CNNs via the softmax activation function. These regularized CNNs not only achieve superior segmentation results but also exhibits enhanced robustness to noise. Based on this work, the authors \cite{33} further introduced non-local TV regularization to the softmax activation function to capture long range dependency information in CNNs. They presented a primal-dual hybrid gradient method for this proposed method. Numerical experiments show that this approach can eliminate isolated regions while preserving more details. Furthermore, instead of using TV regularization, Liu et al. \cite{34} proposed a Soft Threshold Dynamics (STD) framework that can easily integrate various spatial priors, such as spatial regularity, volume constraints and star-shape priori, into CNNs for image segmentation. These methods combine the strengths of both CNNs and model-based methods by applying a variational perspective to the softmax activation function.

%Recently, Jia \textit{et al.} \cite{62} integrated graph total variation into CNNs by augmenting the softmax activation function with a spatial regularity term, enhancing CNNs with spatial coherence. Subsequent work \cite{33} addressed the limitations of convolution operators in capturing long-range dependencies, facilitating the elimination of isolated regions in image segmentation while preserving fine details. Building on this, a new soft threshold dynamic (STD) variational framework is proposed, which integrates spatial regularity, volume constraints and star-shape priori into the Deep Convolutional Neural Networks (DCNNs)\cite{34}. These regularized networks not only achieve superior segmentation results compared to standard CNNs and DCNNs but also exhibits enhanced robustness to noise.
%It is worth mentioning that \cite{45} interprets the classical GNNs layer as a solution to the regularized optimization problem by means of the approximation principle, that is, GNNs is interpretable in the variational perspective. 

Inspired by these works, we aim to apply this variational regularization method to GNNs for node classification tasks, incorporating the non-local TV regularization into the softmax activation function.
It can help GNNs to better capture the spatial information inherent in graphs.
Unlike CNNs, where the similarity relationship between data points must be carefully determined, GNNs inherently provide this relationship through the adjacent matrix. %This approach also mitigates overfitting by incorporating unsupervised regularization, the effectiveness of which in semi-supervised learning has been well demonstrated in \cite{31, 26, 27}. 
We apply the proposed
method into the architecture of GCN and GraphSAGE, testing
them on citation and webpage linking datasets, respectively.
Numerical experiments demonstrate that this variational regularization method is also well-suited for GNNs, and exhibits strong generalization capabilities across 100 random dataset splits and 20 random parameter initializations for each split. 
The contributions of this paper can be summarized as follows:

\begin{itemize}
    \item We introduce a novel graph similarity regularized softmax for GNNs in semi-supervised node classification.
    \item We propose a framework for integrating the traditional variational regularization method into GNNs.
\end{itemize}
% $\bullet$ We introduce a novel graph similarity regularized softmax for GNNs in semi-supervised node classification.

% $\bullet$ We propose a framework for integrating the traditional variational regularization method into GNNs.
%$\bullet$ We present a novel regularized GNNs framework designed for node classification tasks, incorporating the non-local TV regularization into the softmax activation function. It can help GNNs to better capture the spatial information inherent in graphs.
%{\color{red}$\bullet$ We introduce non-local TV regularization into the softmax activation function of common GNN models, enabling them to better capture the inherent spatial information in graphs during the classification process.}

%$\bullet$ We apply this regularized method to GCN and GraphSAGE and employ more robust evaluation procedures based on multiple random splits of train/validation/test sets. The numerical results demonstrate that this approach can achieve good improved accuracy on both assortative and disassortative graphs.
%{\color{red}$\bullet$ We propose a classification framework for GNNs that incorporates variational priors. Within this framework, we maintain the advantages of end-to-end training while alleviating overfitting in GNNs to a certain extent.}

The article is structured as follows: In Section II, we provide a concise overview of related work, including GCN, GraphSAGE and softmax variational problem. Section III presents the regularized GNNs model and algorithm. Experimental results and implementation details are reported in Section IV. We conclude this paper in Section V.
\section{Related Work}
For a graph $\mathcal{G}=\{\mathcal{V},\mathcal{E}\}$, where $\mathcal{V}$ and $\mathcal{E}$ represent the set of nodes and edges, respectively. Let $N$ denote the number of nodes, and let $d$ represent the number of features for each node. A node $x_i\in\mathcal{V}$ has a neighborhood $\mathcal{N}(x_i) =\{x_j\in\mathcal{V}|(x_i, x_j)\in\mathcal{E}\}$, where $(x_i, x_j)$ represents an edge between nodes $x_i$ and $x_j$. The adjacency matrix $A$ is an $N\times N$ matrix with $A_{ij} = 1$ if $(x_i, x_j)\in\mathcal{E}$ and $A_{ij} = 0$ otherwise. The feature matrix $X\in \mathbb{R}^{N\times d}$ represents the features of the nodes, and the nodes are  characterized by $K$ classification labels.

In this section, we first introduce the general form of GNNs and the discuss two special instances: GCN and GraphSAGE. In addition, we give the variational form of softmax function, which lays the foundation for next section.
\subsection{Graph Neural Networks}
%In fact, most GNNs can be considered as some form of message passing, which can be decomposed into two steps: message passing and state updating \cite{51}. 
%{\color{red}Indeed, most GNNs can be considered as some form of Message Passing Neural Networks (MPNNs) \cite{51}. 
%The operation of GNNs is streamlined into two core processes: aggregate and combine, which used for transmitting information from neighbor node and updating the central node's information with aggregated information.}
%This structured approach allows GNNs to effectively process information on graphs. 
The fundamental idea of GNNs is to iteratively update the representation of each node by combing both its own features and the features of its neighboring nodes. In \cite{49}, Xu \textit{et al.} proposed a general framework for GNNs, where each layer consists two important functions as follows:

\begin{itemize}
\item{AGGREGATE function:}
This function aggregates information from the neighboring nodes of each node.
\item{COMBINE function:}
This function updates the node's representation by combining the aggregated information from its neighbors with its current representation.
\end{itemize}

% $\bullet$ AGGREGATE function: This function aggregates information from the neighboring nodes of each node.

% $\bullet$ COMBINE function: This function updates the node's representation by combining the aggregated information from its neighbors with its current representation.

Mathematically, the $l$-th layer of a GNN can be defined as follows:
\begin{equation}
\begin{gathered}
        a_i^{(l)}=\text{AGGREGATE}^{(l)}(\{H_j^{(l-1)}:x_j\in\mathcal{N}(x_i)\}),\\
        i=1,2,...,N,
\end{gathered}
\label{aggr}
\end{equation}
\begin{equation}
    H_i^{(l)}=\text{COMBINE}^{(l)}(H_i^{(l-1)},a^{(l)}_i),i=1,2,...,N,
    \label{comb}
\end{equation}
where $H^{(l)}$ represents the feature matrix at the $l$-th layer (with $H^{(0)}=X$ being the input feature matrix), and $a_i^{(l)}$ denotes the aggregated feature vector of node $x_i$ at the $l$-th layer. %$\mathcal{N}(v)$ denotes the set of sampled neighbor nodes of $v$. 
Different GNNs employ various AGGREGATE and COMBINE functions \cite{49,1,2,3}. 
The node representations in the final layer $H^{(L)}$ can be considered the final representations of the nodes, denoted as  $H^{(L)}=\text{GNN}(X) \in \mathbb{R}^{N \times D}$, then it can be used for node label prediction through the softmax function:
\begin{equation}
\hat{Y} = \text{softmax}(H^{(L)}W).
\end{equation}

where $W \in \mathbb{R}^{D \times K}$ is a learnable matrix.
%For simplicity, we will refer to (\ref{aggr}) and (\ref{comb}) collectively as $\text{GNN}(X)$.
%In this paper, we will uniformly denote the output of all GNNs after $L$ iterations as $H^{(L)}=\text{GNN}(X)$.
\subsection{Graph Convolutional Network}

In \cite{1}, Kipf and Welling proposed a classical GCN. Each layer of GCN can be defined as follows:

\begin{equation}
H^{(l+1)}=\sigma(\tilde{D}^{-\frac{1}{2}}\tilde{A}\tilde{D}^{-\frac{1}{2}}H^{(l)}W^{(l)}),
\end{equation}
where $\tilde{A}=A+I_{N}\in\mathbb{R}^{N\times N}$ is the adjacency matrix of the graph with added self-connections (where $I_N$ is the identity matrix of size $N$), $\tilde{D}$ is the degree matrix of $\tilde{A}$, $W^{(l)}$ is the layer-specific trainable weight matrix, and $\sigma$ is an activation function, such as ReLU and softmax.

%Note that $A$ is adjacency matrix, representing the relationship between nodes, and $\tilde{A}=A+I_{N}\in\mathbb{R}^{N\times N}$ represents adding a self-loop to the adjacency matrix. In the realm of graph-based tasks, the adjacency matrix with an added self-loop is a common consideration. This matrix not only captures the connections and interactions among neighboring nodes but also encapsulates vital information pertinent to the central node itself. $\hat{A}=\tilde{D}^{-\frac{1}{2}}\tilde{A}\tilde{D}^{-\frac{1}{2}}\in\mathbb{R}^{N\times N}$ represents the $ \tilde{A} $ after normalization of the degree matrix $\tilde{D}$, where $\tilde{D}_{ii}=\sum_{j}\tilde{A}_{ij}$ represents the degree matrix of $\tilde{A}$. $H^{l}$ represents the node matrix in the $l$-th layer. $W^{l}$ is the network parameter of the $l$-th layer which is learnable, and $\sigma$ is a nonlinear activation function.

Commonly the feature matrix $X$ and adjacency matrix $A$ of the entire graph are input through a two-layer GCN, defined as follows:

\begin{equation}
\hat{Y}=\text{softmax}(\hat{A}\text{ReLU}(\hat{A}XW^{(0)})W^{(1)}),
\end{equation}
where $\hat{A}=\tilde{D}^{-\frac{1}{2}}\tilde{A}\tilde{D}^{-\frac{1}{2}}\in\mathbb{R}^{N\times N}$  is the normalized adjacency matrix with added self-connections, $W^{(0)}$ and $W^{(1)}$ are the trainable weight matrices of the first and second layers, respectively, and  $\hat{Y}\in\mathbb{R}^{N\times K}$ represents  the output of this network.

GCN is trained by minimizing a cross-entropy loss function, which aims to make the predicted labels of nodes in the training set  as close as possible to their ground truth labels. This loss function is commonly used for classification problems and is given by

\begin{equation}
L(Y,\hat{Y})=-\sum_{i\in \mathcal{Y}_{L}}\sum^{K}_{k=1}Y_{ik}\text{ln}\hat{Y}_{ik},
\label{loss}
\end{equation} 

\noindent
where $\mathcal{Y}_{L}$ is the set of indexes for the nodes used for training. $Y_{i}$ represents the ground-truth probability distribution of the $i$-th node, which is represented as a one-hot vector, with $Y_{ik}$ indicating whether the $i$-th node belongs to the $k$-th class. $\hat{Y}_{ik}$ represents the predicted probability of the $i$-th node belonging to class $k$.

%\subsection{STD}

\subsection{GraphSAGE}
In \cite{3}, Hamilton \textit{et al.} introduced a groundbreaking graph neural network model known as GraphSAGE. %designed for inductive learning. The fundamental concept is that node embeddings can be derived from a universal function that consolidates adjacency information. The key is to train this aggregation function so that it can be generalized to nodes not seen during training.
%The essence of GraphSAGE lies in its Sample and Aggregate (SAmple and AggreGatE) operations. 
During each iteration, a fixed number of neighboring nodes are sampled for each central node, and these nodes are then aggregated as formulated in (\ref{aggr}). 
There are three alternative AGGREGATION functions available:

\begin{itemize}
\item{Mean Aggregator:}
Computes the average of the neighbor embeddings, which is  combined with the central node's embedding through a non-linear trasformation.
\item{LSTM Aggregator:}
Utilizes Long Short-Term Memory (LSTM) networks to process a sequence of neighbor embeddings.
It is important to randomize the order of neighbors to ensure the model does not rely on a specific sequence, as graph structures are inherently unordered.
\item{Pooling Aggregator:}
Applies non-linear transformation followed by pooling operations, such as max or average pooling to the neighbor embeddings to extract the most significant features.
\end{itemize}
% $\bullet$~Mean Aggregator: Computes the average of the neighbor embeddings and combines it with the central node's embedding.
%\begin{equation}
%H_i^{(l)}=\sigma(\mathcal{W}^{(l)}\cdot\text{MEAN}(\{H_i^{(l-1)}\}\cup\{H_j^{(l-1)}:v_j\in\mathcal{N}(v_i)\})),
%\end{equation}

% $\bullet$~LSTM Aggregator: Utilizes Long Short-Term Memory (LSTM) networks to capture the sequence of neighbor embeddings, effectively handling varying orders of node importance.
%\begin{equation}
%\begin{gathered}
%H_i^{(l)}=\sigma(\mathcal{W}^{(l)}\cdot\text{MEAN}(\{H_i^{(l-1)}\}\cup\text{LSTM}_i^{(l-1)}))\\
%\text{LSTM}_i^{(l-1)}=\text{LSTM}\{H_j^{(l-1)}:v_j\in\mathcal{N}(v_i)\}, 
%\end{gathered}
%\end{equation}

% $\bullet$~Pooling Aggregator: Applies non-linear transformation and pooling operations, such as max or average pooling, to the neighbor embeddings to extract the most significant features.
%\begin{equation}
%H_i^{(l)}=\text{MAX}(\{\sigma(H_j^{(l-1)}+\textbf{b}),v_j\in\mathcal{N}(v_i)\}),   
%\end{equation}
%Assuming that GraphSAGE comprises $L$ layers, the aggregation of neighbor information for each central node in each layer can be succinctly described as (\ref{aggr}):

% \begin{equation}
%     % H_v^{(l)} = \text{AGGREGATE}^{(l)}(H_v^{(l-1)}, \{H_u^{(l-1)} | u \in \mathcal{N}(v)\}),
%     \text{AGG}_i^{(l)}=\text{AGGREGATE}^{(l)}(H_j^{(l-1)} | v_j \in \mathcal{N}(v_i)))),
% \end{equation}

%Here, $\text{AGG}_i^{(l)}$ represent the aggregated embedding of the central node $v_i\in\mathcal{V}$'s neighbors at layer $l$. 
%The function $\text{AGGREGATE}^{(l)}$ encapsulates the aggregation process at layer $l$. 
Subsequently, the aggregated information is combined with the state of the central node, as depicted in (\ref{comb}). This COMBINE function is specifically defined by the following formula:

\begin{equation}
      H_i^{(l)} = \sigma(W^{(l)} \cdot \text{CONCAT}(H_i^{(l-1)}, a_i^{(l)})).
\end{equation}
where $\text{CONCAT}$ denotes the concatenation operation and $W^{(l)}$ is the weight matrix at layer $l$. 
By choosing an appropriate aggregation function and training the model with the loss function (\ref{loss}), GraphSAGE can effectively learn node embeddings that generalize well to unseen nodes, making it a powerful tool for inductive learning on graphs.

\subsection{Softmax Variational Problem}

The softmax function is most commonly used in the last layer of neural networks for classification tasks. It converts the input logits into a probability distribution, facilitating the identification of the class with the highest probability as the predicted class.
In \cite{33}, Liu \textit{et al.} provide a  variational interpretation of the softmax function, showing that it can be derived from the following minimization problem:

%The objective of node classification is to categorize nodes into one of $K$ distinct classes. Employing softmax as our classifier, the category with the highest softmax output probability is identified as the predicted class. The softmax classifier is particularly advantageous due to its intrinsic property of being analogous to solving an optimized variational problem, which can be articulated as follows:

\begin{equation}
\begin{gathered}
\min_{\mathscr{A}} -\langle \mathscr{A}, O\rangle+\epsilon\langle \mathscr{A}, \log \mathscr{A}\rangle,\\
\text {s.t.} \sum_{k=1}^{K} \mathscr{A}_{i k}=1, \mathscr{A}_{i k}\geq 0, \forall i=1,2,...,N,
\label{epsilonmin}
\end{gathered}
\end{equation}
where $O=(O_{1},O_{2},...,O_{K})\in \mathbb{R}^{N\times K}$ is the given input, $\mathscr{A}=(\mathscr{A}_{1},\mathscr{A}_{2},...,\mathscr{A}_{K})\in \mathbb{R}^{N\times K}$ is the output we aim to find, the second term of the objective function can be interpreted as a negative entropy term. This term enforces smoothness on $\mathscr{A}$ with a positive control parameter $\epsilon >0$. The equation represents the constraint condition that the sum of each row of $\mathscr{A}$ is 1. Using the Lagrange method, the minimizer of the above problem is:
\begin{equation}
\mathscr{A}^*_{ik}=\frac{\exp\frac{O_{ik}}{\epsilon}}{\sum^{K}_{\hat{k}=1}\exp\frac{O_{i\hat{k}}}{\epsilon}},%i=1,2,...,N,k=1,2,...,K,
\end{equation}
where $O_{ik}$ represents the probability of the $i$-th node belonging to the $k$-th class.
This can be equivalently reformulated in matrix form as
\begin{equation}
\mathscr{A^*}=\text{softmax}\left(\frac{O}{\epsilon}\right).
\end{equation}
In fact, when $\epsilon=1$, it is exactly the softmax activation function, i.e.,

\begin{equation}
	\mathscr{A^*}=\text{softmax}(O).
\end{equation}

\section{Methods and Algorithms}
% \begin{figure}[!t]
%     \centering
%     \includegraphics[width=1.0\linewidth]{model.png}
%     \caption{The architecture of GNLTV}
%     \label{model}
% \end{figure}
\subsection{The Non-local TV Regularized Softmax Function}
Considering that the softmax activation function does not have any spatial regularization, Liu \textit{et al.} \cite{33} proposed a novel regularized softmax activation function by integrating non-local TV regularization, defined as follows:

\begin{equation}
\begin{gathered}
\min_{\mathscr{A}}{\sum_{k=1}^{K}\{-\langle\mathscr{A}_{k},O_{k}\rangle+\epsilon\langle\mathscr{A}_{k},\log\mathscr{A}_{k}\rangle+\lambda NLTV(\mathscr{A}_{k})}\}\\
\text {s.t.}\sum_{k=1}^{K}\mathscr{A}_{ik}=1, \forall i=1,2,...,N,
\end{gathered}
\label{eq_nltv}
\end{equation}

\noindent
where $\lambda$ is a regularization parameter, and the variational formulation of non-local TV is given in the following form

\begin{equation}
NLTV(\mathscr{A}_{k})=
\max _{\|\eta_{k}\|_{\infty}\leqslant 1}\left\langle \mathscr{A}_{k}, \operatorname{div}_{\textbf{S}} \eta_{k}\right\rangle,
\label{2.8}
\end{equation}
where $\eta_k\in \mathbb{R}^{N\times N}$ is the dual variable associated with $\mathscr{A}_k$, $\eta_k(j)$ denotes the $j$-th row of $\eta_k$, and the infinity norm $\|\eta_k\|_{\infty}=\max\limits_{j}\|\eta_{k}(j)\|_{2}$ represents the maximum Euclidean norm among all the rows of the matrix $\eta_k$. By substituting the above equation, the minimization problem (\ref{eq_nltv}) is equivalently reformulated as the following min-max problem:

\begin{equation}
\begin{gathered}
\min_{\mathscr{A}}\max _{\|\eta_{k}\|_{\infty}\leqslant 1}\sum_{k=1}^{K}\{-\langle \mathscr{A}_{k}, O_{k}\rangle+\epsilon\langle \mathscr{A}_{k}, \log \mathscr{A}_{k}\rangle\\+\lambda\left\langle \mathscr{A}_{k}, \operatorname{div}_{\textbf{S}} \eta_{k}\right\rangle\},\\
\text {s.t.}\sum_{k=1}^{K} \mathscr{A}_{ik}=1, \forall i=1,2,...,N.
\label{2.9}
\end{gathered}
\end{equation}

By employing the alternating minimization algorithm to solve this problem, it can be decomposed into two subproblems for $\eta$ and  $\mathscr{A}$  respectively:

\begin{itemize}
\item{\textbf{$\eta$-subproblem:}}

For fixed $\mathscr{A}$, we solve
\begin{equation}
\max _{\|\eta_{k}\|_{\infty}\leqslant 1}\left\langle \mathscr{A}_{k}, \operatorname{div}_{\textbf{S}} \eta_{k}\right\rangle.
\label{2.11}
\end{equation}
\end{itemize}
% \textbf{$\bullet$~$\eta$-subproblem:}

% For fixed $\mathscr{A}$, we solve
% \begin{equation}
% \max _{\|\eta_{k}\|_{\infty}\leqslant 1}\left\langle \mathscr{A}_{k}, \operatorname{div}_{\textbf{S}} \eta_{k}\right\rangle.
% \label{2.11}
% \end{equation}

\begin{itemize}
\item{\textbf{$\mathscr{A}$-subproblem:}}

For fixed $\eta$, we solve
\begin{equation}
\begin{gathered}
\min_{\mathscr{A}}\sum_{k=1}^{K}\{-\langle\mathscr{A}_{k},O_{k}\rangle+\epsilon\langle\mathscr{A}_{k},\log\mathscr{A}_{k}\rangle+\lambda\left\langle\mathscr{A}_{k},\operatorname{div}_{\textbf{S}}\eta_{k}\right\rangle\}\\
 \text {s.t.}\sum_{k=1}^{K}\mathscr{A}_{ik}=1. %\forall i=1,2,...,N.
	\label{2.10}
\end{gathered}
\end{equation}
\end{itemize}
% \textbf{$\bullet$~$\mathscr{A}$-subproblem:} 

% For fixed $\eta$, we solve
% \begin{equation}
% \begin{gathered}
% 	\min_{\mathscr{A}}\sum_{k=1}^{K}\{-\langle \mathscr{A}_{k}, O_{k}\rangle+\epsilon\langle \mathscr{A}_{k}, \log \mathscr{A}_{k}\rangle+\lambda\left\langle \mathscr{A}_{k}, \operatorname{div}_{\textbf{S}} \eta_{k}\right\rangle\},\\
%  \text {s.t.}\sum_{k=1}^{K} \mathscr{A}_{ik}=1, %\forall i=1,2,...,N.
% 	\label{2.10}
% \end{gathered}
% \end{equation}

For the $\eta$-subproblem, it is solved using the Lagrange multiplier technique, while for the $\mathscr{A}$-subproblem, it is tackled through gradient descent combined with a projection operator. 
More detailed information on solving these two subproblems can be found in Appendix (\ref{A}).
Then, the iteration can be expressed as:

\begin{equation}
	\begin{cases}
	\begin{aligned}
	%\eta^{t}_{k}&=\prod\nolimits_{\left\|\eta_{k}\right\|_{\infty}\leqslant 1} \left(\eta^{t-1}_{k}-\tau \nabla_{w} \mathscr{A}_{k}^{t-1}\right),\\
    \eta^{t}_{k}&=\mathcal{P}_B \left(\eta^{t-1}_{k}-\tau \nabla_{\textbf{S}} \mathscr{A}_{k}^{t-1}\right),\\
	\mathscr{A}^{t}_{k}&=\operatorname{softmax}\left(\frac{O_{k}-\lambda\operatorname{div}_{\textbf{S}} \eta^{t}_{k}}{\epsilon}\right),
	\end{aligned}
	\end{cases}
	k=1,2,...,K,
	\label{2.15}
\end{equation}

\noindent
where $\bigtriangledown_{\textbf{S}}\mathscr{A}_{k}\in \mathbb{R}^{N\times N}$ and $\operatorname{div}_{\textbf{S}}\eta_{k}\in \mathbb{R}^{N}$ represent the non-local gradient operator and non-local divergence operator, respectively.

The non-local gradient operator ${\bigtriangledown}_{\textbf{S}}u(x):L^{2}(\mathcal{V})\rightarrow L^{2}(\mathcal{V}\times \mathcal{V})$ is defined as
\begin{equation}
({\bigtriangledown}_{\textbf{S}}u)(x_{i},x_{j})=\textbf{S}(x_{i},x_{j})(u(x_{j})-u(x_{i})), i=1,2,...,N.
\end{equation}

The non-local divergence operator $\operatorname{div}_{\textbf{S}}v(x):L^{2}(\mathcal{V}\times \mathcal{V})\rightarrow L^{2}(\mathcal{V})$ is defined as
\begin{equation}
\begin{gathered}
(\operatorname{div}_{\textbf{S}}v)(x_{i})=\sum _{x_{j}\in \mathcal{N}(x_{i})} \textbf{S}(x_{i},x_{j})(v(x_{i},x_{j})-v(x_{j},x_{i})),\\
i=1,2,...,N.
\end{gathered}
\end{equation}
\noindent
Here, $\textbf{S}(x_i,x_j) \geq 0$ is the weight function that measures the similarity between two nodes $x_i$ and $x_j$. In \cite{33}, Liu \textit{et al.} provide several methods to define a proper weight function to depict the relationship between data points, which are crucial for grid-structured data like images. However, for graph-structured data, the relationship between nodes have already been directly given in the form of adjacency matrix $A$. To ensure stable training, we use the normalized form to define the similarity matrix $\textbf{S}=D^{-\frac{1}{2}}AD^{-\frac{1}{2}}$, with $D$ being the diagonal degree matrix of $A$.

%{\color{red}The similarity matrix $ \textbf{w} $ quantifies the similarity between pairs of nodes, thereby enabling the effective utilization of the pairwise relationships present in graph data. Specifically, $\textbf{w}_{ij}\in\mathbb{R}$ represents the similarity between $i$-th node and $j$-th node. In the research by Liu \textit{et al.}, the weight matrix $ \textbf{w} $ is determined through a variety of unique definitions \cite{33}. However, in GCN, the similarity between nodes can be represented by the adjacency matrix of the graph structure. Therefore, to strengthen numerical stability throughout the compution process, we define $\textbf{w}=D^{-\frac{1}{2}}AD^{-\frac{1}{2}}$, where $D_{ii}=\sum_{j}A_{ij}$ means the degree matrix of $A$.}

The projection operator is given by

%\begin{equation}
%\Pi_{\left\|\xi_{k}\right\|_{\infty}\leq 1}(\xi_{k})=
%\begin{cases}
%\begin{aligned}
%&y,& 
%\left\|y\right\|_{2} \leq 1,\\
%\end{aligned} \\
%\begin{aligned}
%&\frac{y}{\left\|\xi_{k}\right\|_{\infty}},& 
%\left\|y\right\|_{2}>1,
%\end{aligned}
%\end{cases}
%\end{equation}
%where $y$ is the $j$-th line of %$\xi_{k}$. 

\begin{equation}
\mathcal{P}_B(\eta_{k})=
\begin{cases}
\begin{aligned}
&\eta_k(j),& 
\left\|\eta_k(j)\right\|_{2} \leq 1,\\
%\end{aligned} \\
%\begin{aligned}
&\frac{\eta_k(j)}{\left\|\eta_{k}\right\|_{\infty}},& 
\left\|\eta_k(j)\right\|_{2}>1,
\end{aligned}
\end{cases}
\end{equation}
where $\eta_k(j) \in \mathbb{R}^N$ denotes the $j$-th row of $\eta_{k}$ and $\|\eta_k\|_{\infty}=\max\limits_{j}\|\eta_{k}(j)\|_{2}$. 

Therefore, the non-local TV regularized softmax function can be expressed as:

\begin{equation}
	\mathscr{A}^{*}=\operatorname{softmax}\left(\frac{O-\lambda\operatorname{div}_{\textbf{S}} \eta}{\epsilon}\right).
\end{equation}

\subsection{The Proposed RGNN Algorithm}

In this section, we propose to apply the regularized non-local TV softmax function within the GNN framework. 
In the forward process, we first initialize the learnable weight matrices $W^{(l)},l=1,...,L$ in the GNN models. 
For each epoch, we then initialize $\mathscr{A}^{0}$ as follows:
\begin{equation}
\mathscr{A}^{0}=\text{softmax}(O),
\end{equation}
where $O=\text{GNN}(X)\in \mathbb{R}^{N\times K}$ represents the output node features of the GNN before applying the softmax function. Here, $\mathscr{A}^{0}$ corresponds to the output classes generated by the forward process of traditional GNN models. Based on these results and setting $\eta^{(0)}=\textbf{0}$, we proceed to compute the following iterations to update $\eta$ and $\mathscr{A}$ for the non-local TV regularized softmax function:

%\subsubsection{ $\eta$-subproblem}
\begin{itemize}
\item{\textbf{$\eta$-subproblem:}}

\textbf{Step 1.} Compute the non-local gradient operator: \\
\begin{equation}
\begin{gathered}
	(\bigtriangledown_{\textbf{S}}\mathscr{A}^{t-1}_{k})(x_{i},x_{j})=\textbf{S}(x_{i},x_{j})(\mathscr{A}^{t-1}_{jk}-\mathscr{A}^{t-1}_{ik}),\\
 i, j=1,2,...,N, k=1,2,...,K,
\label{compute-grad}
\end{gathered}
\end{equation}
\noindent
where the similarity matrix $\textbf{S}=D^{-\frac{1}{2}}AD^{-\frac{1}{2}}$. In the context of the graph structure, the selection of $\textbf{S}$ is linked to the adjacency matrix $A$ of the given graph.  

\noindent
\textbf{Step 2.} Update $\eta_k$ using the gradient descent method and the projection operator:

\begin{equation}
\eta_{k}^t=\mathcal{P}_B(\eta^{t-1}_{k}-\tau \bigtriangledown_{\textbf{S}}\mathscr{A}^{t-1}_{k}), k=1,2,...,K.\\
\label{compute-grad-descent}
\end{equation}
\end{itemize}
% \textbf{$\bullet$~$\eta$-subproblem:}\\
% Step 1. Compute the non-local gradient operator:\\
% \begin{equation}
% \begin{gathered}
% 	(\bigtriangledown_{\textbf{S}}\mathscr{A}^{t-1}_{k})(x_{i},x_{j})=\textbf{S}(x_{i},x_{j})(\mathscr{A}^{t-1}_{jk}-\mathscr{A}^{t-1}_{ik}),\\
%  i, j=1,2,...,N, k=1,2,...,K,
% \label{compute-grad}
% \end{gathered}
% \end{equation}
% \noindent
% where the similarity matrix $\textbf{S}=D^{-\frac{1}{2}}AD^{-\frac{1}{2}}$. In the context of the graph structure, the selection of $\textbf{S}$ is linked to the adjacency matrix $A$ of the given graph.  

% \noindent
% Step 2. Update $\eta_k$ using the gradient descent method and the projection operator:

% \begin{equation}
% \eta_{k}^t=\mathcal{P}_B(\eta^{t-1}_{k}-\tau \bigtriangledown_{\textbf{S}}\mathscr{A}^{t-1}_{k}), k=1,2,...,K.\\
% \label{compute-grad-descent}
% \end{equation}

%Step 2: compute gradient descent

%\begin{equation}
%\eta_{k}=\eta^{t-1}_{k}-\tau \bigtriangledown_{w}\mathscr{A}^{t-1}_{k}, k=1, \ldots ,K,\\
%\label{compute-grad-descent}
%\end{equation}

%Step 3: compute the projection operator
%\begin{equation}
%\eta_{k}^{t}=\Pi_{\left\|\eta_{k}\right\|_{\infty}\leq 1}(\eta_{k})=
%\begin{cases}
%\begin{aligned}
%&y,& 
%\left\|y\right\|_{2} \leq 1,\\
%&\frac{y}{\left\|\eta_{k}\right\|_{\infty}},& 
%\left\|y\right\|_{2}>1,
%\end{aligned}
%\end{cases}
%\label{compute-pro}
%\end{equation}
%where $y$ is the j-th line of $\eta_k$, j=1, $\ldots$ ,N.

%\subsubsection{$\mathscr{A}$-subproblem}
\begin{itemize}
\item{\textbf{$\mathscr{A}$-subproblem:}}

\textbf{Step 1.} Compute the non-local divergence operator:
\begin{equation}
\begin{gathered}
(\operatorname{div}_{\textbf{S}}\eta^{t}_{k})(x_{i})=\sum_{x_{j}\in \mathcal{N}(x_{i})}\textbf{S}(x_{i},x_{j})(\eta^{t}_{k}(x_{i},x_{j})\\
-\eta^{t}_{k}(x_{j},x_{i})),\\
i=1,2,...,N, k=1,2,...,K.
\label{compute-div}
\end{gathered}
\end{equation}

\noindent
\textbf{Step 2.} Calculate the non-local TV regularized softmax function:

\begin{equation}
	\mathscr{A}^{t}_{k}=\text{softmax}\left(\frac{O_{k}-\lambda \operatorname{div}_{\textbf{S}}\eta^{t}_{k}}{\epsilon}\right), k=1,2,...,K.
	\label{compute-softmax}
\end{equation}
\end{itemize}
% \textbf{$\bullet$~$\mathscr{A}$-subproblem:}\\
% Step 1. Compute the non-local divergence operator:
% \begin{equation}
% \begin{gathered}
% (\operatorname{div}_{\textbf{S}}\eta^{t}_{k})(x_{i})=\sum_{x_{j}\in \mathcal{N}(x_{i})}\textbf{S}(x_{i},x_{j})(\eta^{t}_{k}(x_{i},x_{j})\\
% -\eta^{t}_{k}(x_{j},x_{i})),\\
% i=1,2,...,N, k=1,2,...,K.
% \label{compute-div}
% \end{gathered}
% \end{equation}

% \noindent
% Step 2. Calculate the non-local TV regularized softmax function:

% \begin{equation}
% 	\mathscr{A}^{t}_{k}=\text{softmax}\left(\frac{O_{k}-\lambda \operatorname{div}_{\textbf{S}}\eta^{t}_{k}}{\epsilon}\right), k=1,2,...,K.
% 	\label{compute-softmax}
% \end{equation}

After iterating to obtain the optimal matrix $\hat{Y}=\mathscr{A}^T$, we proceed to compute the cross-entropy loss $L(\hat{Y},Y)$ between $\hat{Y}$ and the ground-truth probability distribution matrix $Y$.
In the backward process, we compute the gradient of the loss $L$ with respect the parameters $W^{(l)}$ using automatic differentiation. This process traverses the computational graph backward and is commonly implemented in machine learning frameworks. 
This gradient $\frac{\partial L}{\partial W^{(l)}}$ is then used in popular optimizers like stochastic gradient descent (SGD) \cite{52} or its variants to update the parameters $W^{(l)}$ during training.

\begin{figure}[ht]%[htbp]
	\centering
\subfloat[original GNN]{\includegraphics[scale=0.25]{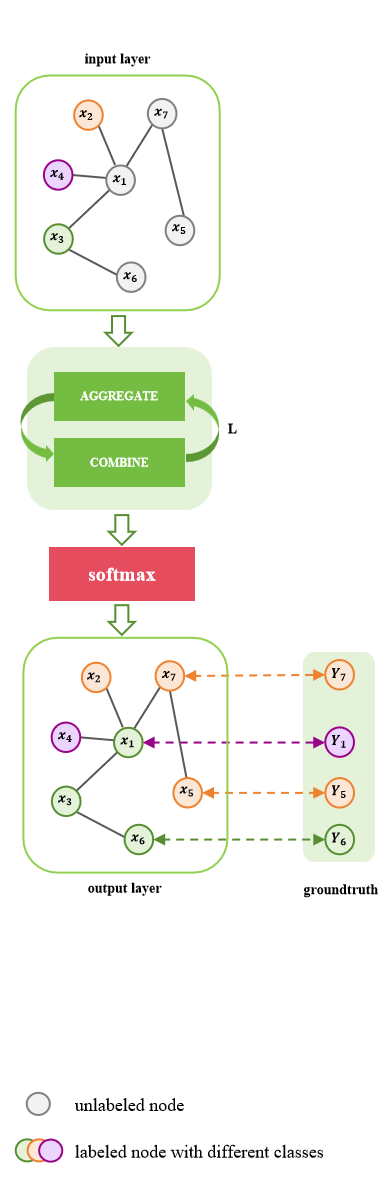}
\label{modelgnn}}
\hfil
\subfloat[RGNN]{\includegraphics[scale=0.25]{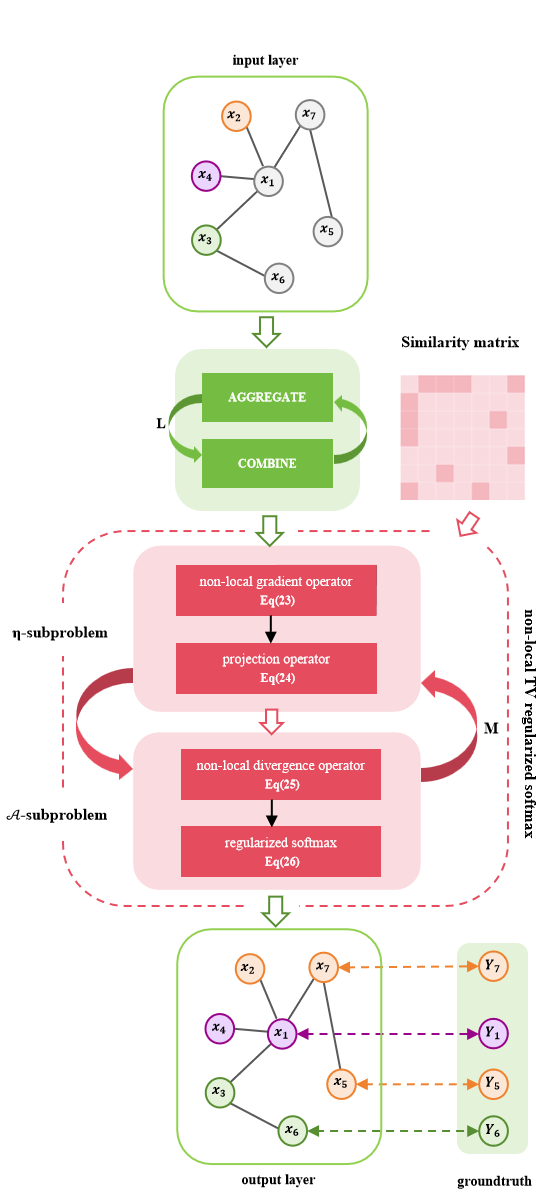}
\label{modelrgnn}}
\caption{GNN vs. RGNN: A structural comparison. (The output shows that node $x_1$ is misclassified by the GNN but correctly classified by the RGNN.)}
\label{fig:model}	
\end{figure}

We summarize the proposed RGNN algorithm in Algorithm \ref{1}. To better understanding, Figure \ref{fig:model} illustrates the differences between the structures of GNN and RGNN.
	\begin{algorithm}
		\LinesNumbered
		\KwIn{Graph $\mathcal{G}(\mathcal{V},\mathcal{E})$,  feature matrix $X$, learnable parameters $\lambda, \epsilon, \tau$,
                external iterations number $M$ and internal iterations number $T$.			
		}
		
		\textbf{Initialize:} 
            $W^{(l)},l=1,2,...,L$.\\
		\For{$epoch=1,2,...,M$}{
			$O=\text{GNN}(X)$, $\mathscr{A}^{0}=\text{softmax}(O)$,\\
			\textbf{Set} $\eta^0=\textbf{0}$.\\
                \For{$t=1,2,...,T$}{
				\textbf{$\eta$-subproblem:}\\
				\qquad (1)compute the non-local gradient operator $\bigtriangledown_{\textbf{S}}\mathscr{A}^{t-1}$ by (\ref{compute-grad}).\\
				\qquad (2)update $\eta^t$ using gradient descent and projection operator by (\ref{compute-grad-descent}).\\
				
				\textbf{$\mathscr{A}$-subproblem:}\\
				\qquad (1)compute the non-local divergence operator $\operatorname{div}_{\textbf{S}}\eta^{t}$ by (\ref{compute-div}).\\
				\qquad (2)calculate the regularized softmax function $\mathscr{A}^t$ by (\ref{compute-softmax}).}
                Get the matrix $\hat{Y}=\mathscr{A}^T$ and compute the cross-entropy function $L(Y, \hat{Y})$.\\	
			Update $W^{(l)}$ by using
                gradient descent method.
               }             
        \KwOut{$W^{(l)}, l=1,2,..,L$.}
	\caption{RGNN algorithm}  
	\label{1}
	\end{algorithm}
%To enhance clarity, a comparison of RGNN and GNN structures is presented in Figure \ref{fig:model}.

\section{Experiments}
\subsection{Datasets}
In this paper, we discuss the performance of our proposed RGNN approach on both assortative and disassortative datasets.
\subsubsection{Assortative datasets}
We utilize three standard citation network benchmark datasets for the semi-supervised node classification task: Cora, Citeseer, and Pubmed \cite{46,47}. In these datasets, nodes represent individual documents, while edges represent citation relationships between these documents. Node features are extracted from the bag-of-words representation of the document content, and each node is associate with a corresponding class label. 
\subsubsection{Disassortative datasets} We employ three specialized webpage linking network benchmark datasets: Cornell, Texas, Wisconsin, which are subdatasets of WebKB \cite{54}, collected from computer science departments of various universities. In these datasets, nodes represent web pages and edges are hyperlinks between them. Node features are bag-of-words representation of web pages. 

We provide the statistics of each dataset in Table \ref{dataset}. %The label rate in Table \ref{dataset} denotes the proportion of labeled nodes used for training relative to the total number of nodes in each dataset.

% \begin{table*}[h]
% 	\centering
% 	\caption{Graph Dataset Statistics.}
% 	\label{dataset}
%         \tabcolsep=0.7cm
% 	\begin{tabular}{ccccccc}
% 		\hline
% 		\textbf{Dataset} & \textbf{Assortativity} & \textbf{Type} & \textbf{Classes} & \textbf{Features} & \textbf{Nodes} & \textbf{Edges} \\ \hline
% 		Cora             & assortative &Citation network      & 7              & 1433              & 2485           & 5069           %& 0.056
%   \\
% 		Citeseer         & assortative & Citation network     & 6              & 3703              & 2120           & 3679           %& 0.054
%   \\
% 		Pubmed           & assortative & Citation network     & 3              & 500               & 19717          & 44324          %& 0.003           
%          %& 0.6 
%   \\ \hline
%               Cornell          & disassortative & Internet network   & 5           & 1703              & 183            & 295 
%   \\ 
%             Texas            & disassortative & Internet network   & 5           & 1703              & 183            & 309 
%   \\ 
%             Wisconsin        & disassortative & Internet network   & 5           & 1703              & 251            & 499 
%   \\ \hline
% 	\end{tabular}
% \end{table*}
\begin{table}[ht]
\caption{Graph Dataset Statistics.}
\label{dataset}
\tabcolsep=0.3cm
\begin{tabular}{cccc|ccc}
\hline
         & \multicolumn{3}{c|}{\textbf{Assortative}} & \multicolumn{3}{c}{\textbf{Disassortative}} \\
\textbf{Datasets}  & Cora    & Citeseer    & Pubmed   & Cornell    & Texas   & Wisconsin   \\ \hline
\textbf{Classes}  & 7       & 6           & 3        & 5          & 5       & 5           \\
\textbf{Features} & 1433    & 3703        & 500      & 1703       & 1703    & 1703        \\
\textbf{Nodes}    & 2485    & 2120        & 19717    & 183        & 183     & 251         \\
\textbf{Edges}    & 5069    & 3679        & 44324    & 295        & 309     & 499         \\ \hline
\end{tabular}
\end{table}
\subsection{Parameters initialization}

All experiments are conducted using PyTorch on an NVIDIA RTX 3090 GPU with 24GB of memory.
%Our computation platform, a Linux server with a 24GB RTX 3090 GPU, supports our experiments conducted using Python 3.8.13 and PyTorch 1.10.0+cu111. 
We train our proposed RGNN models for a maximum of 10000 epochs, with early stopping implemented if the validation loss does not decrease or the accuracy does not increase for 50 consecutive epochs. %To ensure fair comparisons, the size of hidden layer units in our proposed regularized model remains consistent with those used in GCN \cite{1}. 
All the network parameters are initialized using Glorot initialization and optimized using the Adam optimizer \cite{48}. 

Since the optimal hyperparameters may vary across different random splits and initializations, we treat these three hyperparameters $ \tau $, $ \lambda $ and $ \epsilon $ in the proposed RGNN model as learnable, starting with initial values. 
Table \ref{para_setting} outlines our parameters initialization and learning rate. To optimize computation time, our experiments generally focus on the case where $T=1$. Despite this limitation, this regularized softmax approach still proves effective in our numerical experiments. %For the STD-GCN model, we set the hyperparameter $\epsilon$ to 3 across all datasets. The number of iterations $T$ is set to $10$ for the Cora dataset and $1$ for the Citeseer and Pubmed datasets.

\begin{table}[H]
	\centering
	\caption{Initial values and learning rate (Lr) of parameters for RGNN.}
	\label{para_setting}
	\begin{tabular}{cccccccc}
		\hline
		& Cor. & Cit. & Pub. & Cor. & Tex. & Wis. & Lr\\ \hline
		$\tau_{ini}$& 1.0 & 0.3 & 1.0 & 0.01 & 0.01 & 0.01 & 0.01\\
		$\lambda_{ini}$& 3.0 & 8.0 & 3.0 & 0.3 & 3.0 & 1.0 & 0.001\\ 
		$\epsilon_{ini}$& 1.0 & 5.0 & 0.5 & 8.0 & 3.0 & 1.0 & 0.01\\ \hline
	\end{tabular}
\end{table}

\subsection{Experimental Results}

In this section, we conduct a thorough comparison of our proposed RGNN method with several prominent graph learning approaches: GCN \cite{1}, GAT \cite{2}, and two variants of GraphSAGE (mean and maxpool) \cite{3}. We evaluate the average classification performance of each method across 2000 experiments, with 100 random dataset splits and 20 random initializations. % are reported in Table \ref{results}. We can observe that our regularized GCN models demonstrate remarkable competitiveness across these experiments. Our NLTV-GCN model achieves the highest performance on the Cora dataset, whereas our STD-GCN model demonstrates superior performance on the Citeseer and Pubmed datasets. Both the NLTV-GCN and STD-GCN models consistently rank among the top performers across all 2000 experiments on various datasets.

\subsubsection{Experiments on assortative datasets}
%We explore the semi-supervised node classification challenge within a GNNs framework as GCN, utilizing well-known graph datasets such as Cora, Citeseer, and PubMed. Our approach involves enhancing the GCN model with an improved methodology. These datasets, which are citation networks of considerable size and uniformity. Specifically, the experiments were conducted with 100 distinct random splits and 20 varied initializations. 
Recognizing the superior performance of GCN on assortative datasets-Cora, Citeseer, and Pubmed-we integrate the non-local TV regularized softmax into GCN, resulting in the regularized GCN (RGCN). We compare the performance of RGCN with standard GCN, GAT and GraphSAGE-mean and GraphSAGE-maxpool.
For each dataset split, the training set is formed by randomly selecting 20 nodes per class, the validation set by selecting 30 nodes per class, with the remainder assigned to the test set. The results, presented in Table \ref{results}, indicate that the RGCN model outperforms the standard GCN and consistently achieves the highest average accuracy across all these three citation datasets. This highlights the efficiency of the softmax regularization and its robust generalization capabilities.

\begin{table}[htbp]
	\centering
	\caption{The mean accuracy and std (\%) over 100 random dataset splits and 20 random initializations for each split.}
	\label{results}
	\begin{tabular}{cccc}
		\hline
		& Cora              & Citeseer          & Pubmed             \\ \hline
		GCN \cite{1}                    & 81.5±1.3          & 72.1±1.6          & 79.0±2.1\\
		GAT \cite{2}                    & 80.8±1.3          & 71.6±1.7          & 78.6±2.1\\
		GraphSAGE-mean \cite{3}        & 79.3±1.3          & 71.6±1.6          & 76.1±2.0\\
		GraphSAGE-maxpool \cite{3}          & 76.7±1.8          & 67.4±2.2          & 76.9±2.0\\ \hline
		RGCN(ours)         & \textbf{81.9±1.1}          & \textbf{74.1±1.6}          & \textbf{79.2±2.1}     \\ \hline
	\end{tabular}
\end{table}

\subsubsection{Experiments on disassortative datasets}
%To further evaluate the model's scalability, we extended our analysis to the GraphSAGE setting by integrating the enhanced technique into the original model. %WebKB, a dataset comprising web pages from renowned computer science departments, nodes as web pages, edges as hyperlinks, and features attributes as bag-of-words representations. It encompasses three distinct sub-datasets: Cornell, Texas, and Wisconsin. 

%\textbf{GCNs rely on the assortativity assumption, where similar nodes tend to be interconnected. Experiments, as documented in \cite{55}, indicate that GCN, along with GAT, excel in academic networks characterized by strong assortativity. In contrast, GraphSAGE's inductive learning approach, which constructs an embedding for a target node by sampling and aggregating neighbor features, is structurally not sensitive and theoretically more adept at generalizing to unseen nodes.} 

Considering GraphSAGE's proficiency with disassortative datasets, we integrate the non-local TV regularized softmax into GraphSAGE-mean, resulting in RGraphSAGE-mean. We compare its performance with standard GraphSAGE-mean, GraphSAGE-maxpool, GCN, and GAT on Cornell, Texas, and Wisconsin datasets.
Due to the small number of nodes in some classes, we modified the partitioning strategy for dataset splits, randomly allocating 60\%, 20\%, and 20\% of the nodes to the training, validation, and test sets, respectively. 
%Employing the mean accuracy of 100 distinct random splits and 20 varied initializations, we established a benchmark for performance assessment. %Figure \ref{sp} and 
Table \ref{spresults} demonstrates that RGraphSAGE-mean significantly outperforms GraphSAGE-mean and other test methods across these webpage datasets, underscoring its practical efficacy of non-local TV regularized softmax. Additionally, these average results over 2000 experiments suggest that our regularized model maintains robust performance across different dataset splits and initializations.

\begin{table}[htbp]
	\centering
	\caption{The mean accuracy and std (\%) over 100 random dataset splits and 20 random initializations for each split.}
	\label{spresults}
	\begin{tabular}{cccc}
		\hline
		& Cornell              & Texas          & Wisconsin             \\ \hline
		GCN \cite{1}                  & 36.7±3.9          & 42.4±3.3          & 51.8±3.3\\
		GAT \cite{2}                  & 41.2±3.3          & 46.8±5.0          & 51.8±4.2\\
		GraphSAGE-mean \cite{3}   & 76.4±6.0          & 78.6±4.1          & 78.7±2.1\\
		GraphSAGE-maxpool \cite{3}    & 66.6±5.8          & 77.5±4.2          & 74.0±1.6\\ \hline
		RGraphSAGE-mean(ours)         & \textbf{80.8±6.4}          & \textbf{81.4±6.2}          & \textbf{80.8±5.7}     \\ \hline
	\end{tabular}
\end{table}
% \begin{figure}[!t]
%     \centering
%     \includegraphics[width=1.0\linewidth]{vis/newdata.png}
%     \caption{Experimental results for node classification on three common social network datasets.}
%     \label{sp}
% \end{figure}
\subsection{Further Discussion}
\subsubsection{Comparison of time consumption}
%Considering the computational efficiency, the incorporation of the non-local TV block enriches the node representations by integrating non-local global information, leading to a moderate increase in the model's training duration. 
By incorporating the non-local TV regularized softmax into GNNs, addition computation is required compared to original GNNs, which can increase training time.
Table \ref{time} and Table \ref{time1} provide a comparison of time expenditure for RGCN, RGraphSAGE-mean and their counterparts during a single training epoch.
It can be observed that the training time has a certain increase compare to GCN and GraphSAGE. However, this additional computational cost can lead to significant improvement in model accuracy and generalization.
\begin{table}[htbp]
	\centering
	\caption{Time(s) per epoch for different GNNs on citation datasets.}
	\label{time}
\begin{tabular}{cccc}
\hline
               & Cora     & Citeseer & Pubmed \\ \hline
GCN \cite{1}            & 3.55e-3 & 3.72e-3 & 3.69e-3  \\
GAT \cite{2}           & 8.78e-3 & 8.97e-3 & 1.53e-2  \\
GraphSAGE-mean \cite{3} & 4.69e-3 & 4.34e-3 & 5.10e-3 \\
GraphSAGE-maxpool \cite{3}  & 4.45e-3 & 6.05e-3 & 8.16e-3 \\ \hline
RGCN(ours)          & 2.98e-3 & 2.87e-2 & 2.71e-1 \\ \hline
\end{tabular}
\end{table}

\begin{table}[htbp]
    \centering
    \caption{Time(s) per epoch for different GNNs on webpage datasets.}
    \label{time1}
\begin{tabular}{cccc}
\hline
               & Cornell  & Texas    & Wisconsin \\ \hline
GCN \cite{1}            & 3.41e-3 & 3.51e-3 & 3.64e-3  \\
GAT \cite{2}            & 8.72e-3 & 8.13e-3 & 8.50e-3  \\
GraphSAGE-mean \cite{3} & 4.65e-3 & 4.59e-3 & 4.74e-3  \\
GraphSAGE-maxpool \cite{3}  & 2.79e-3 & 2.85e-3 & 2.96e-3  \\ \hline
RGraphSAGE-mean(ours)          & 4.05e-2 & 3.85e-2 & 3.79e-2  \\ \hline
\end{tabular} 
\end{table}

\subsubsection{Parameters analysis}
In this section, we discuss the performance of the RGCN model with different initializations for the three parameters  $\tau$, $\lambda$, and $\epsilon$. Specifically, $\tau$ serves as the learning rate for the gradient descent in the $\eta$-subproblem, $\lambda$ acts as the coefficient for the non-local TV regularization term, and $\epsilon$ represents the coefficient for the negative entropy term in the RGCN model. 
%\subsection{Parameters analysis}

\begin{figure}[htbp]%[H]
	\centering
        \subfloat[$\tau$]{\includegraphics[width=0.48\linewidth]{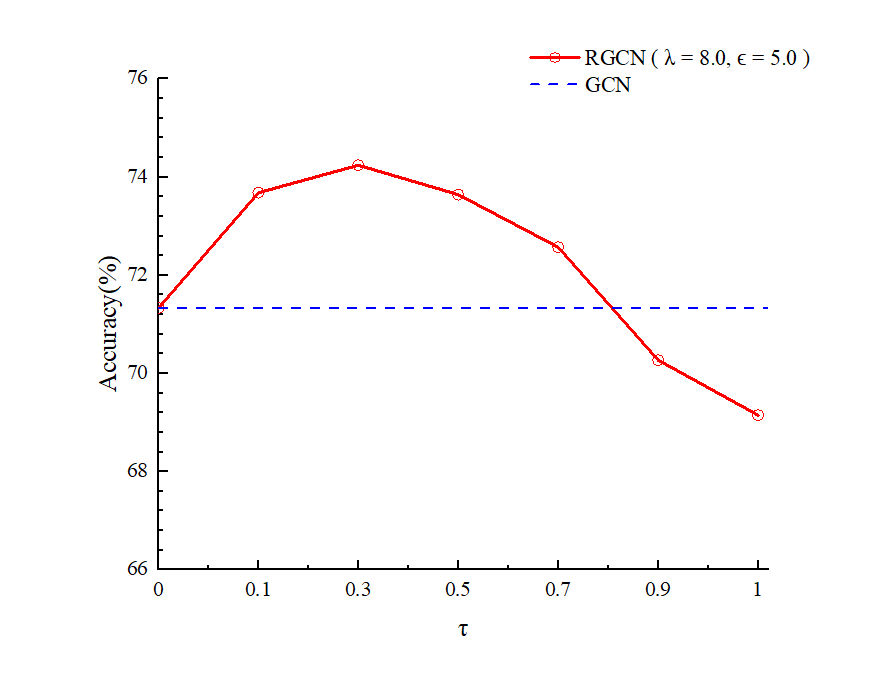}
        \label{para_a}
        }
        \hfil
        \subfloat[$\lambda$]{\includegraphics[width=0.48\linewidth]{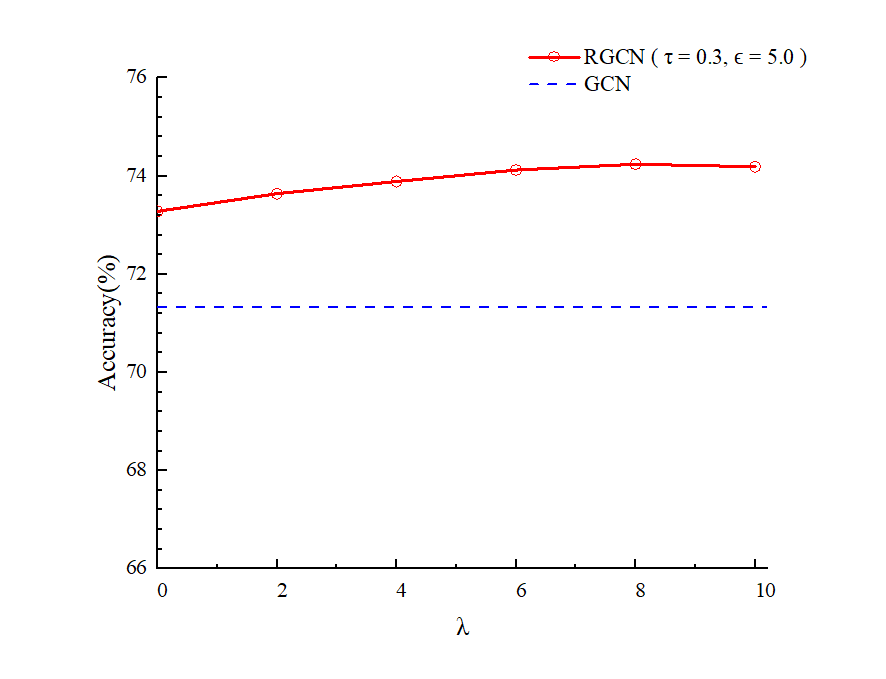}
        \label{para_b}
        }
        \hfil
        \subfloat[$\epsilon$]{\includegraphics[width=0.48\linewidth]{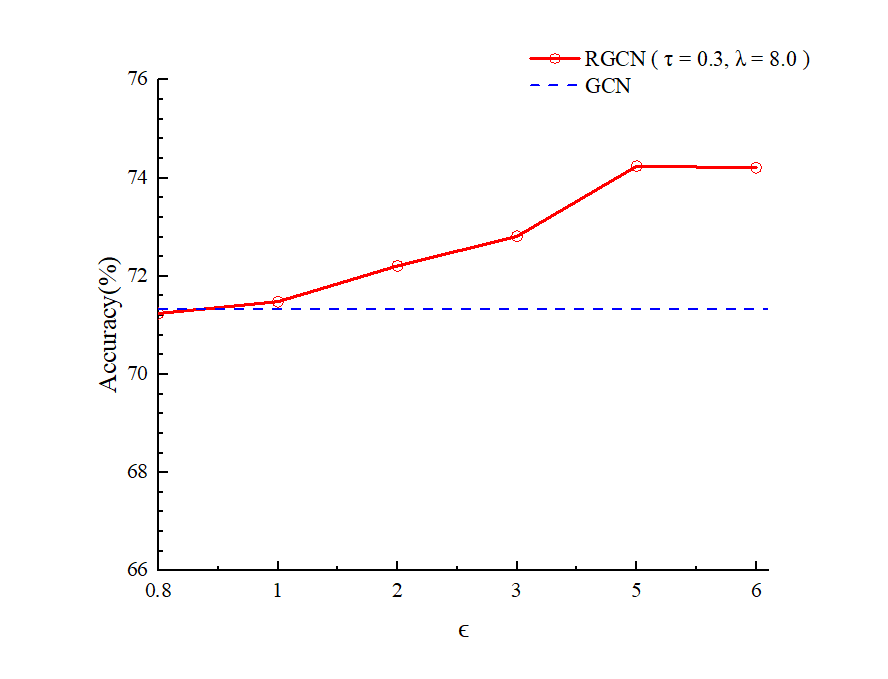}
        \label{para_c}
        }
	\caption{Classification accuracy on Citeseer dataset of the RGCN model with varying initial settings of parameters $\tau$,$\lambda$ and $\epsilon$, respectively.}
	\label{fig:para_ana}	
\end{figure}

In Figure \ref{fig:para_ana}, 
we evaluate the influence of each hyperparameter by keeping the initial values of the other two hyperparameters fixed. 
The average classification accuracy are obtained from 5 random data splits and 5 random parameter initializations on Citeseer dataset.
We observe that when the initial parameter values are within a certain range, the proposed RGCN model (represented by the red solid line) can consistently outperforms the baseline GCN model (indicated by the blue dashed line), highlighting the advantages of this non-local TV regularized softmax in GCN for semi-supervised node classification. 

In Figure \ref{fig:para_ana}(a), with initial values set to $ \lambda=8.0 $ and $ \epsilon=5.0 $, it is evident that within the range of $ (0,0.7] $ for parameter $ \tau $, the RGCN model achieves superior accuracy compared to the standard GCN model, reaching its highest accuracy at $\tau =0.3$. 
%When delving into specifics, the $ \tau $ parameter ranges from $0$ to $0.3$, the experimental results exhibit an upward trend. Conversely, when it ranges from $0.3$ to $1$, the experimental results demonstrate a downward trend. 
In addition, we discuss the parameter $ \lambda $ within $ [0,10] $ with initial values set to $\epsilon=5.0$ and $\tau =0.3$. As depicted in Figure \ref{fig:para_ana}(b), experimental results of RGCN consistently outperform the GCN within this range, with the best performance achieved at $\lambda=8.0$. 
Similarly, the parameter $\epsilon$ is also very important for the experimental results shown in Figure \ref{fig:para_ana}(c), where the highest accuracy is achieved at $\epsilon = 5.0$.

\begin{figure}[ht]%[H]
    \centering
    \includegraphics[width=1.0\linewidth]{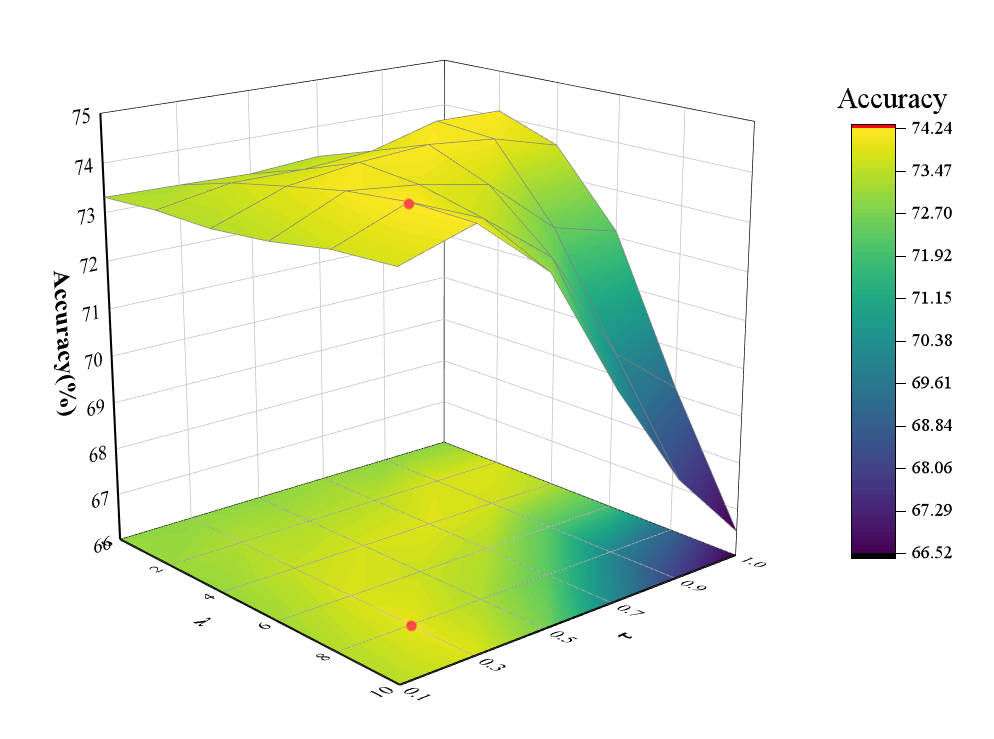}
    \caption{The combined impact of the scaling factor $\tau$ and regularization parameter $\lambda$ on Citeseer dataset.}
    \label{taolambda}
\end{figure}

%In the RGCN model, $\tau$, $\lambda$, and $\epsilon$ are three core parameters. $\tau$ and $\lambda$ serve as the key hyperparameters of the non-local TV module, while $\epsilon$ governs the scaling technique. 
Furthermore, we conducted an in-depth study on how the parameter $\tau$ and $\lambda$ collectively influence performance of RGCN on the Citeseer dataset, with the initial value of $\epsilon=5.0$. As depicted in Figure \ref{taolambda}, when the initial value of $\tau$ approaches $1.0$, the model tends to perform poorly, especially as $\lambda$ also increases. Conversely, when the initial value of $\tau$ is closer to $0.1$, the model is more likely to exhibit superior performance. This indicates a significant correlation between these two parameters, with the optimal combination being $(\tau, \lambda) = (0.3, 8.0)$, achieving the highest accuracy of $74.24\%$, as shown in Figure \ref{taolambda}. 
Additionally, Figure \ref{fig:para_ana}(c) highlights that the intial value of $\epsilon$ is crucial for the performance of the RGNN model when $\tau$ and $\lambda$ are set to 0.3 and 8.0, respectively.
These results indicate that the careful tuning of these three hyperparameters is essential for optimizing model performance, and proper combination of $\tau$, $\lambda$, and 
$\epsilon$ will enable the proposed RGNN model to achieve its best performance.
%During the parameter tuning processes, we first explore the appropriate order of magnitude within the range of $0.1$ to $10.0$, and then further refine our search to identify the optimal values among $\{1, 3, 5, 8\}$.
\subsubsection{Visualization}

\begin{figure*}[tp]%[!t]%[htbp]
\centering
\subfloat{\includegraphics[width=0.32\linewidth]{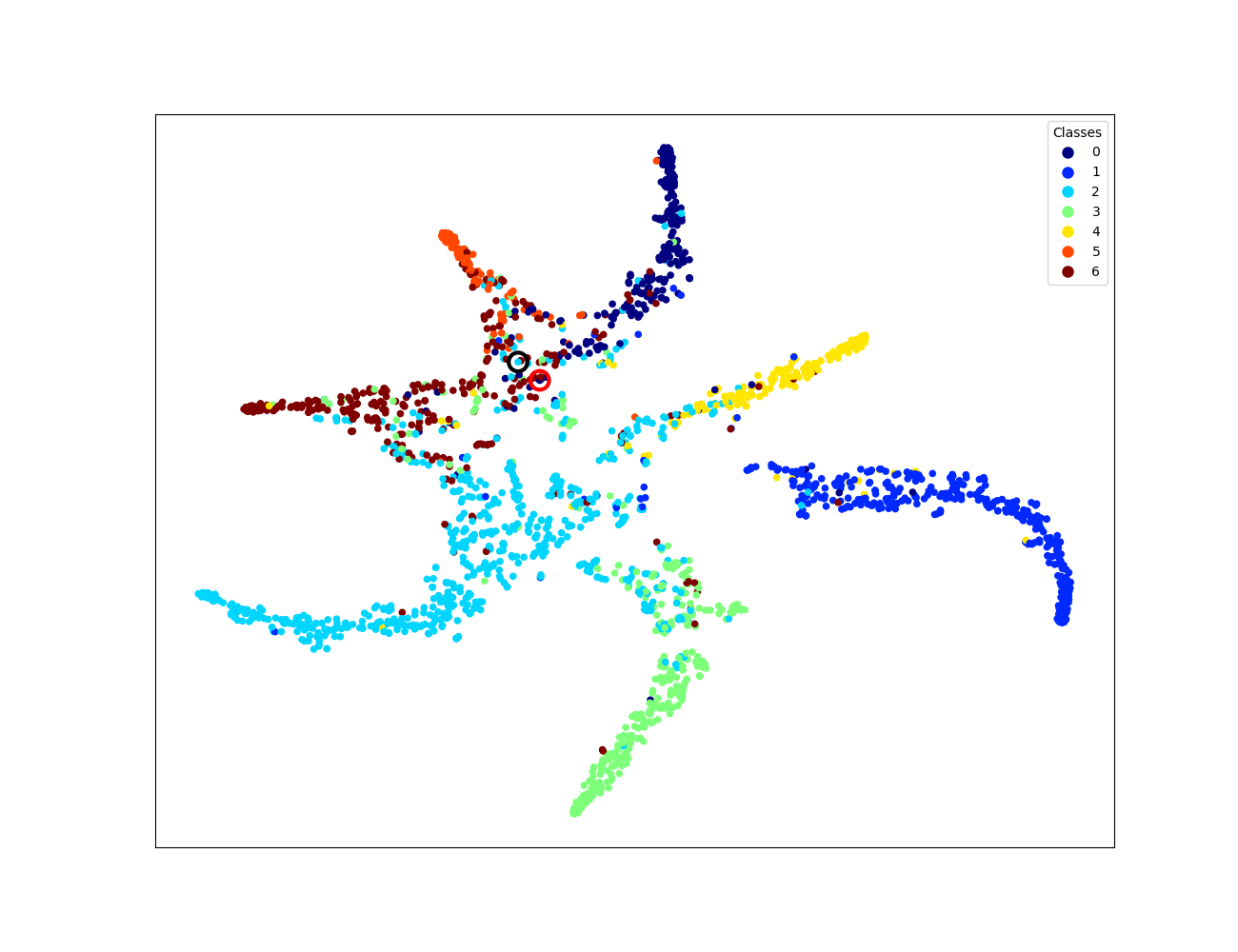}}
\hfil
\addtocounter{subfigure}{-1}
\subfloat[GCN]{\includegraphics[width=0.32\linewidth]{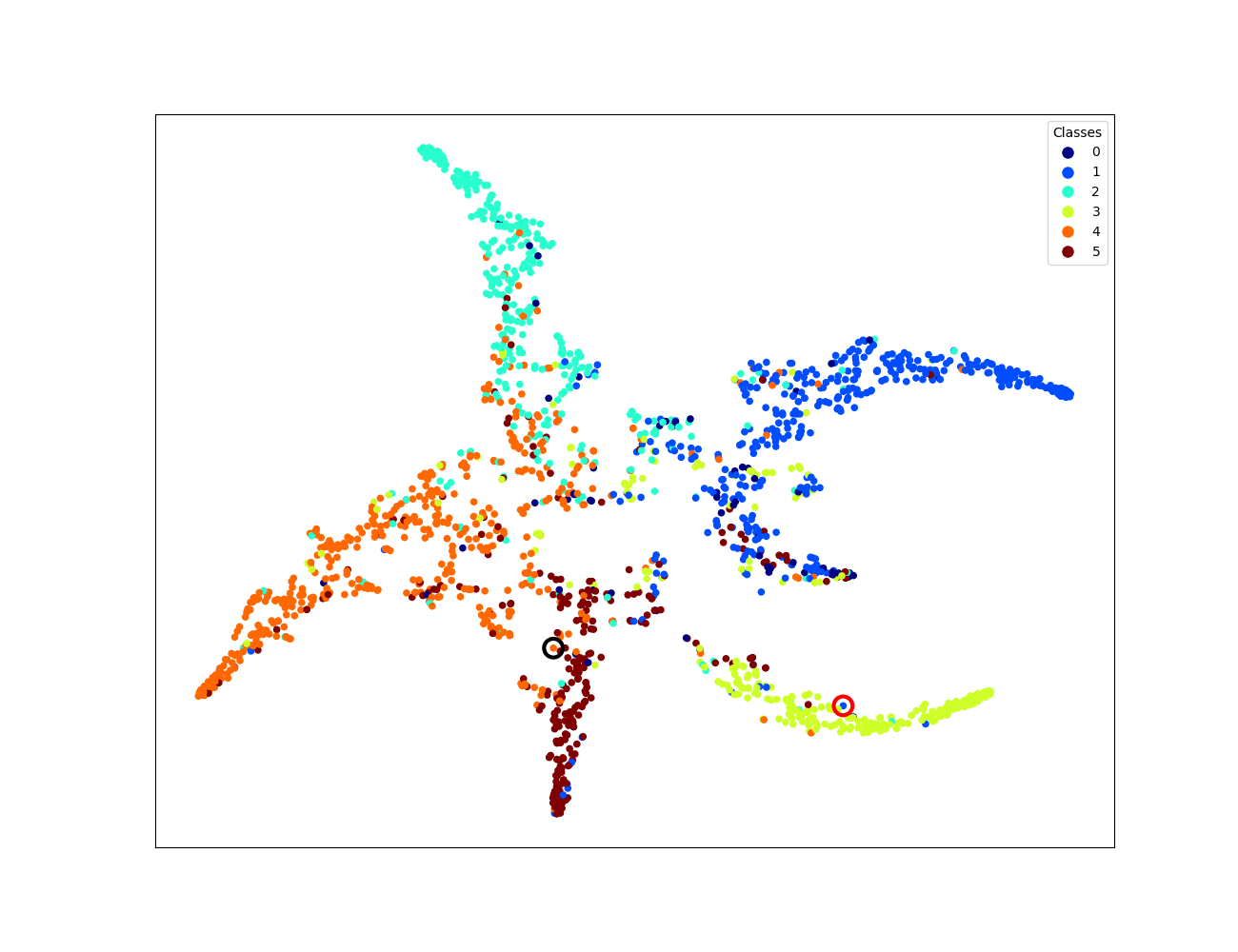}
\label{visa}}
\hfil
\subfloat{\includegraphics[width=0.32\linewidth]{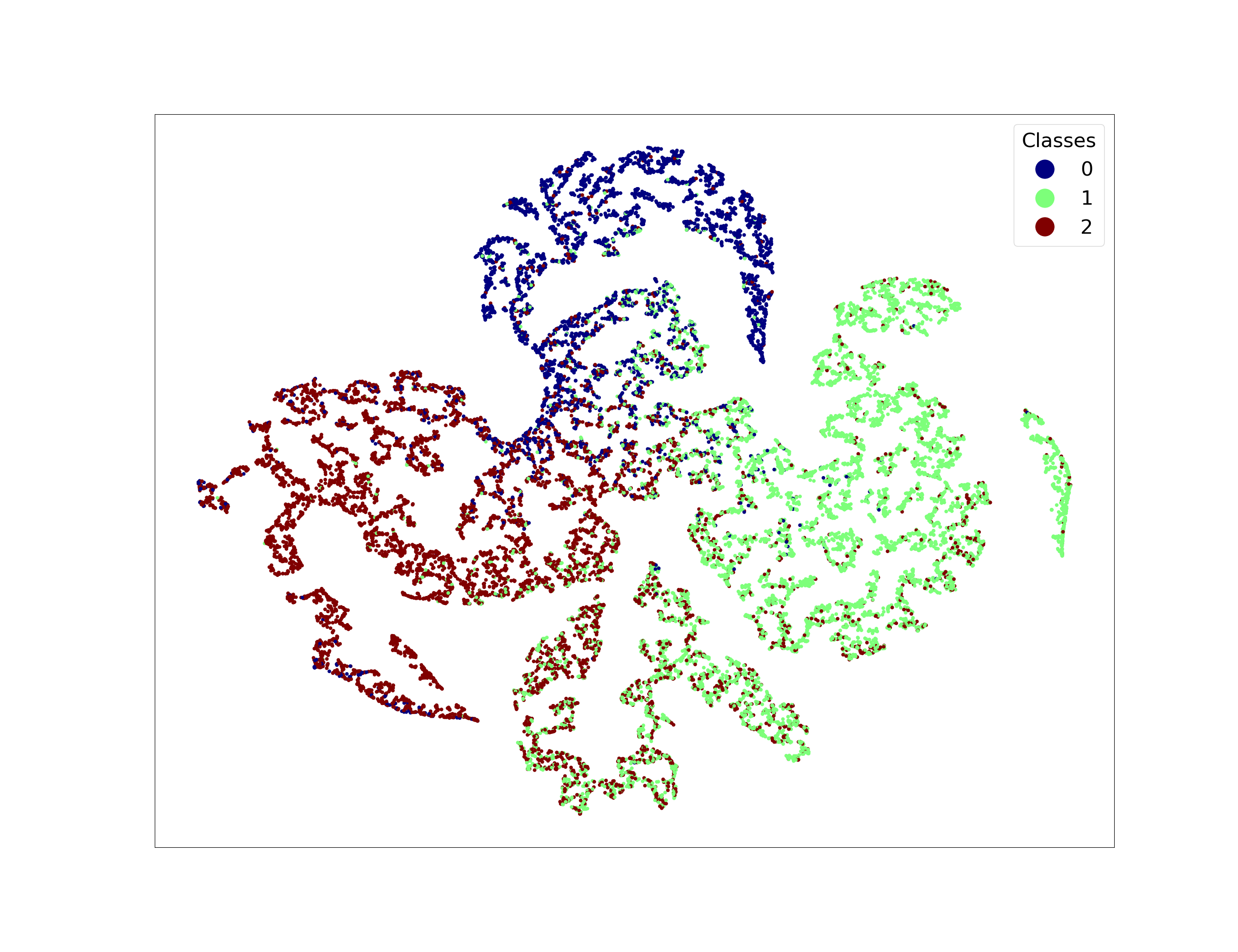}}
\hfil
\subfloat{\includegraphics[width=0.32\linewidth]{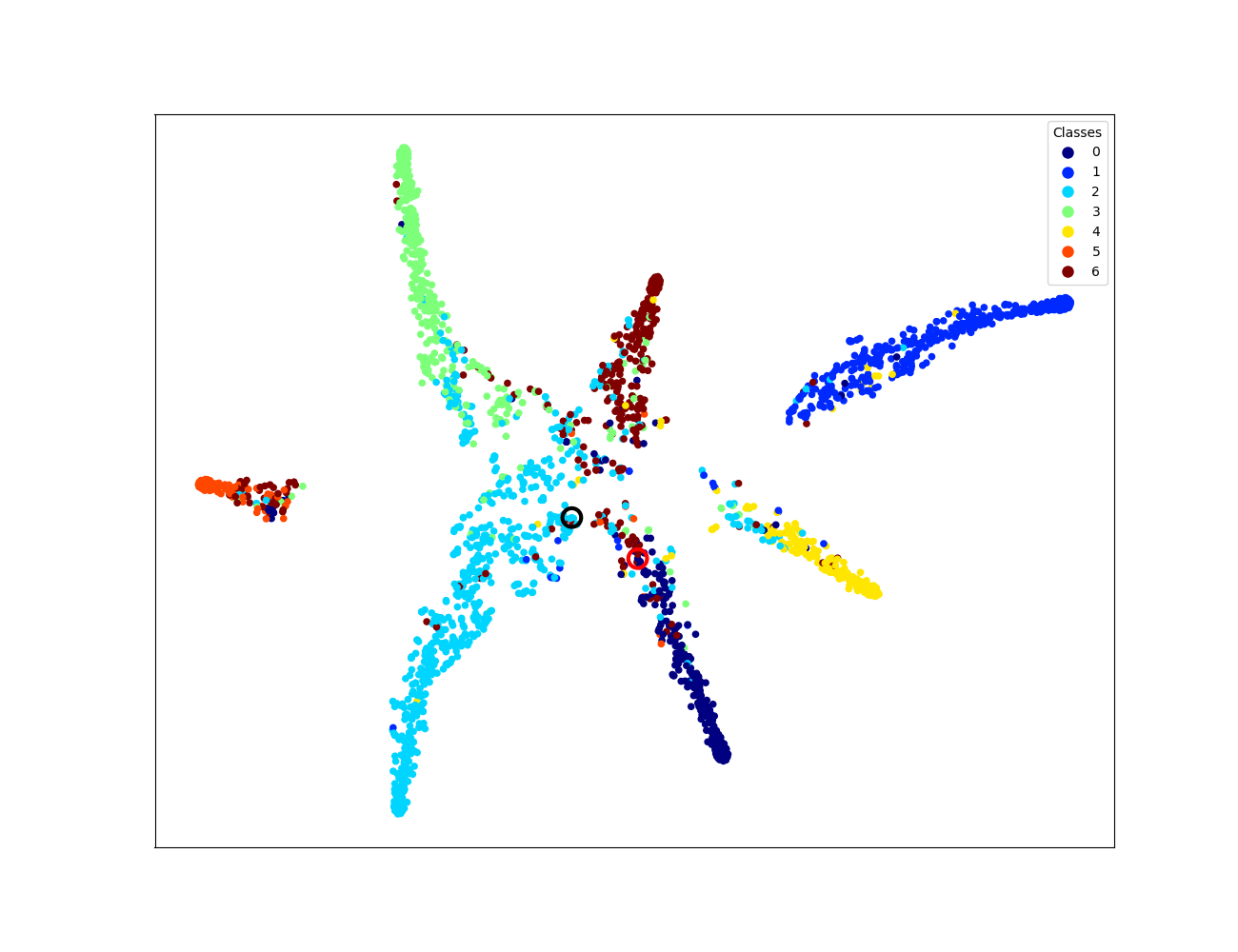}}
\hfil
\addtocounter{subfigure}{-2} 
\subfloat[RGCN]{\includegraphics[width=0.32\linewidth]{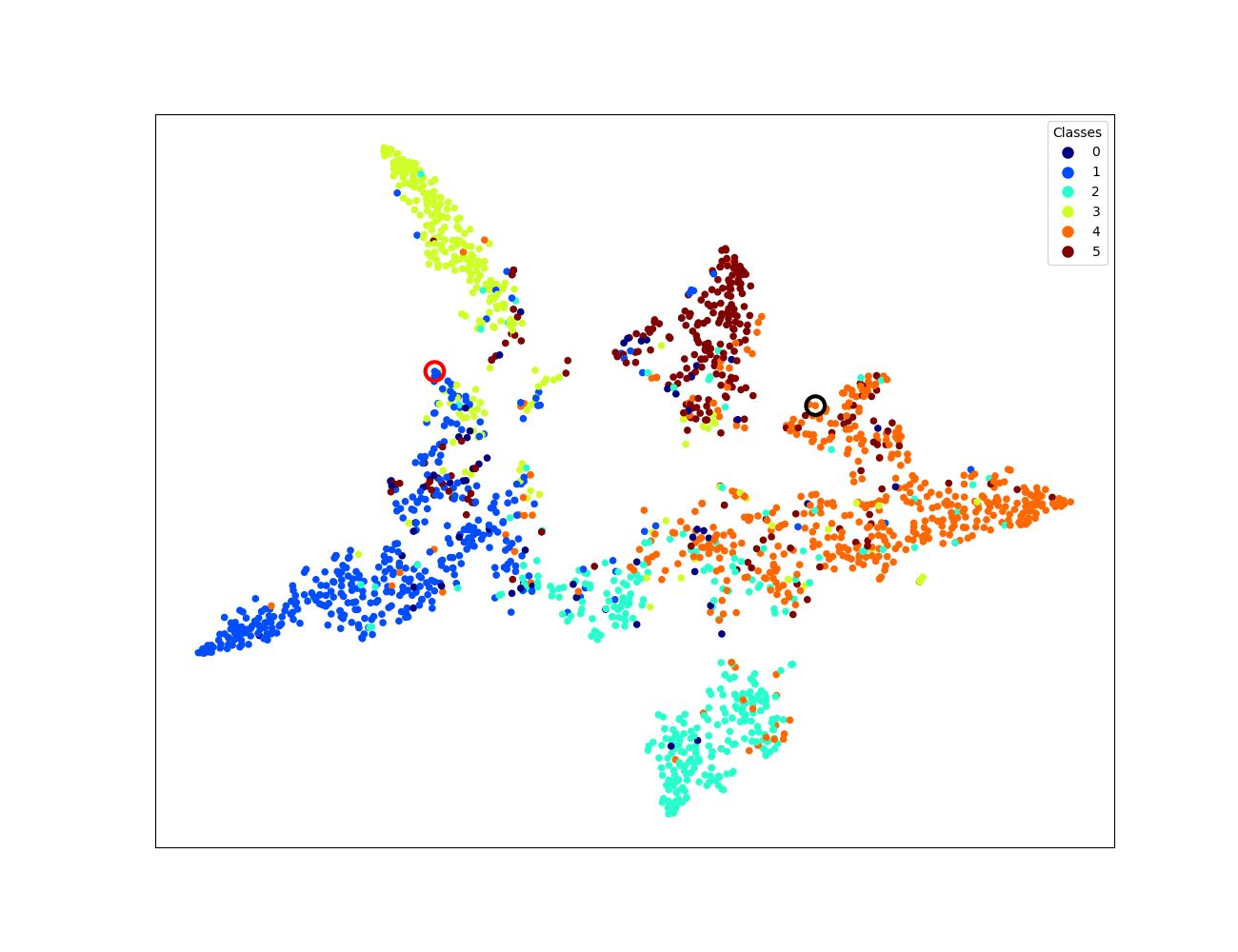}
\label{visb}}
\hfil
\subfloat{\includegraphics[width=0.32\linewidth]{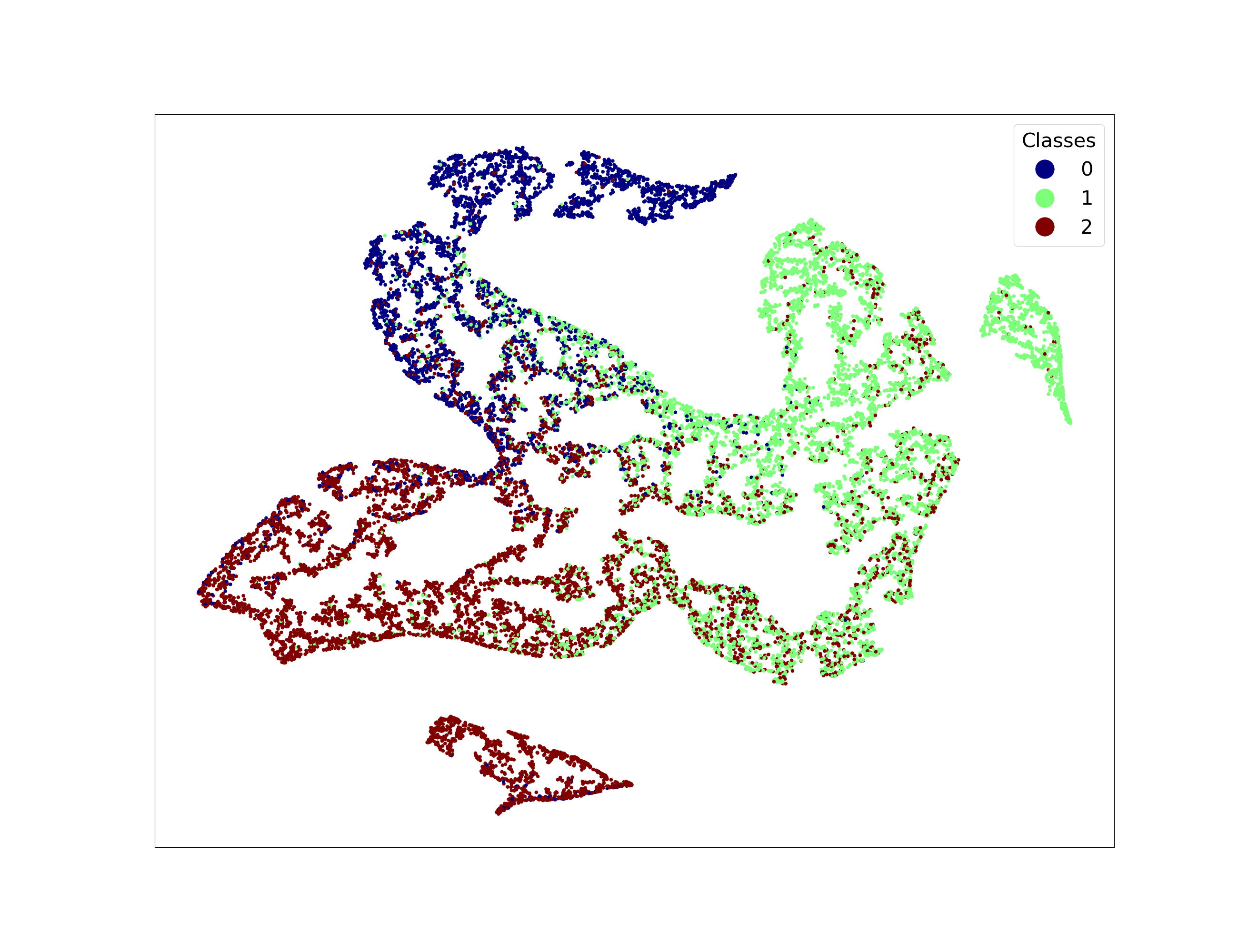}}
% \hfil
% \subfloat{\includegraphics[width=0.7\linewidth]{vis/cmap.png}}
\caption{Visualization results of GCN and RGCN on Cora, Citeseer and Pubmed datasets (from left to right).}
\label{vis}
\end{figure*}

To intuitively illustrate the classification performance of our regularized GNN models, we utilize the t-Distributed Stochastic Neighbor Embedding (t-SNE) technique \cite{41} to visualize the predict results of GCN after applying the standard softmax function and RGCN after applying the regularized softmax function. The visualization of classification results for Cora, Citeseer and Pubmed datasets are presented in Figure \ref{vis}. In these visualizations, each data point represents an individual node from the test sets, with distinct colors denoting the ground truth classes of the nodes. The spatial positions of nodes reflect the projections of high-dimensional embeddings generated by t-SNE.
If nodes from different classes are well-separated and nodes from the same class are grouped closely together, it indicates that the model has higher prediction accuracy and better effectiveness.

The outputs generated by RGCN, as shown in Figure \ref{vis}(b), are more separable than those produced by GCN, as shown in Figure \ref{vis}(a). For instance, class 2 and class 4 in Cora dataset, and class 4 and class 5 in Citeseer dataset. Several nodes that were misclassified by GCN are indicated with red and black circles; these nodes have been correctly reclassified by RGCN. For example, in Core dataset, a node within a black circle was incorrectly classified by GCN as class 5 but was correctly assigned to class 2 by RGCN. Similarly, in Citeseer dataset, a node within a red circle was misclassified by GCN as class 3 but was accurately reclassified by RGCN as class 1. These results confirm that the regularized softmax function is more effective than the standard softmax function in GNNs for semi-supervised node classification.

\subsubsection{Ideal similarity matrix}

\begin{figure}[ht]%[htbp]
	\centering
\subfloat[A given graph]{\includegraphics[width=0.25\textwidth]{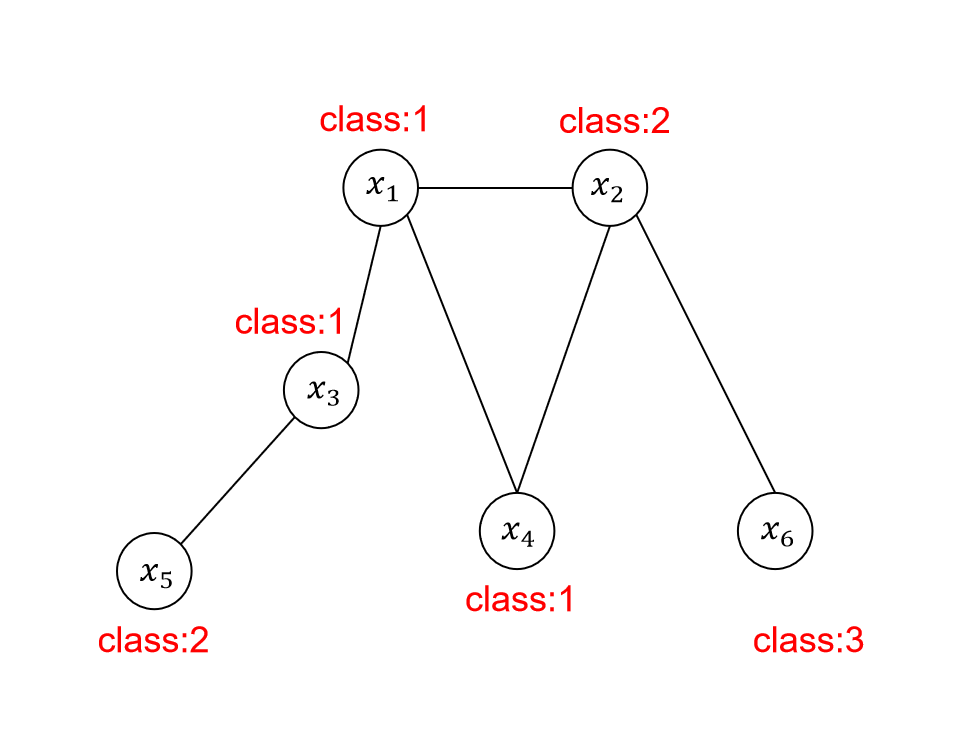}
\label{11}}
\hfil
\subfloat[Ideal $\textbf{S}$]{\includegraphics[width=0.2\textwidth]{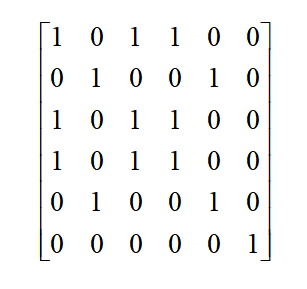}
\label{22}}
\caption{An example of constructing an ideal similarity matrix $\textbf{S}$. }
\label{fig:idealizedw}	
\end{figure}

%In addition, when we think about the improvement of non-local regularization algorithm, we also think about whether we can improve the similarity matrix $w$ of graphs. Then, we need to verify the effect of $w$ in the algorithm. In the process, we attempt a method similar to \textit{cheating}. 
In this section, to demonstrate the importance of choosing the similarity matrix $\textbf{S}$ in our models, we design an ideal similarity matrix. We assume that the label of each node on the given graph is known and can be used to construct the ideal similarity matrix according to the following rule: if nodes $x_i$ and $x_j$ belong to the same class, then $\textbf{S}(x_i,x_j)=1$; if nodes $x_i$ and $x_j$ belong to different classes, then $\textbf{S}(x_i,x_j)=0$. For better understanding, we provide an example of constructing the ideal similarity matrix from a given graph, as shown in the Figure \ref{fig:idealizedw}. 

\begin{figure}[ht]%[htbp]
\centering
\subfloat[Cora]{\includegraphics[width=0.5\linewidth]{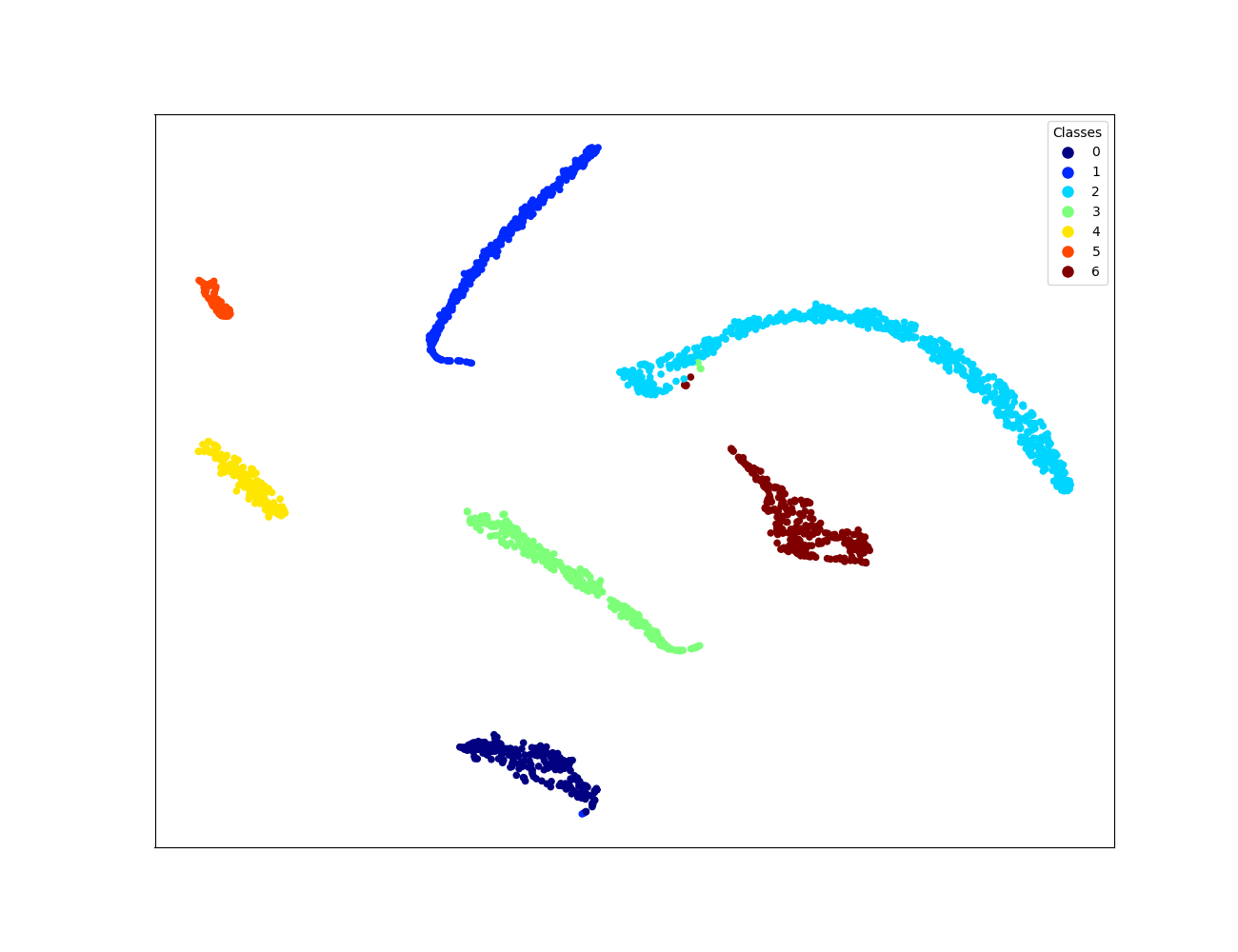}}
\hfil
\subfloat[Citeseer]{\includegraphics[width=0.5\linewidth]{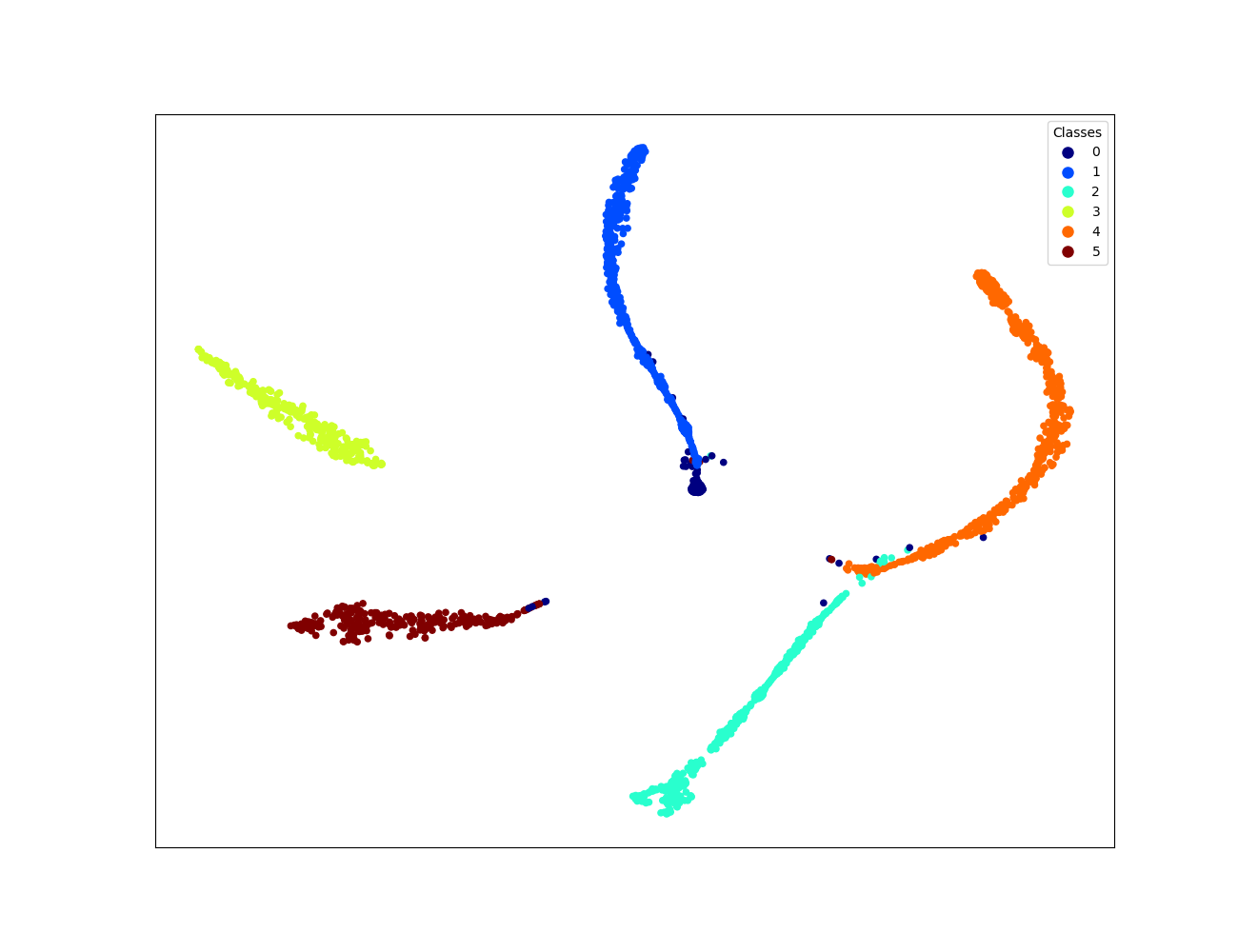}}
\caption{A visualization for the outputs obtained from RGCN with ideal similarity matrix $ \textbf{S} $ on the Cora and Citeseer datasets. }
\label{fig:idealed_w}	
\end{figure}

%\subfloat[]{}
%In Figure \ref{11}, we present a graph depicting multiple nodes, each with its known label indicated directly on the graph. Figure \ref{22}, on the other hand, illustrates the similarity matrix constructed based on the labels of the nodes. 
In Table \ref{tab:ideal_w}, we present the results of GCN, RGCN and RGCN using the ideal similarity matrix $\textbf{S}$ on a single random dataset split and initialization. We can clearly observe that the classification accuracy achieved with ideal $\textbf{S}$ is notably high, approaching nearly 100\%, surpassing both GCN and RGCN.
Additionally, the visualizations of RGCN with ideal similarity matrix $\textbf{S}$ on the Cora and Citeseer datasets are displayed in Figure \ref{fig:idealed_w}.
The results show outstanding performance with clear separation between nodes with different labels and high cohesion among nodes sharing the same label. The excellent experimental results further validate the effectiveness of our method, demonstrating that a well-constructed similarity matrix $\textbf{S}$ can significantly enhance the model's performance.

%Examining Table \ref{tab:ideal_w} and Figure \ref{fig:idealed_w}, it becomes evident that the node classification excels under the idealized similarity matrix, showcasing exceptional performance that nodes bearing distinct labels are sharply segregated, and the cohesion among nodes sharing the same label is exceptionally high. This also intuitively proves that if $\textbf{w}$ is constructed well, it can have an exciting impact on the effect of the model.

\begin{table}[ht]
	\centering
	\caption{The accuracy (\%) over a single random dataset split and initialization.}
	\label{tab:ideal_w}
	\begin{tabular}{ccc}
		\hline
		& Cora & Citeseer \\ \hline
		GCN \cite{1}                  & 79.8 & 73.0     \\
		RGCN(ours)             & 81.8 & 73.7     \\
		RGCN-ideal $\textbf{S}$ & \textbf{95.1} & \textbf{95.0}     \\ \hline
	\end{tabular}
\end{table}

\section{Conclusion}
This paper presents a non-local TV  regularized softmax for GNNs aimed at semi-supervised node classification tasks.
In our experiments, we apply the proposed regularization method to both GCN and GraphSAGE and observe better performance on assortative and disassortative datasets. To demonstrate the generalization capability of our approach, we conduct experiments over 100 train/validation/test splits and 20 random initializations for each dataset.
Additionally, we discuss the significant impact of the similarity matrix on model performance. We propose that designing more effective similarity matrix, rather than relying solely on the adjacency matrix, is a promising direction for future research. For example, developing a learnable similarity matrix could be beneficial for this task. Furthermore, we will explore incorporating STD regularization into the GNN framework.

\printbibliography
\appendix
\section*{Calculating the $\mathscr{A}$-subproblem and $\eta$-subproblem in RGNN}
\label{A}
%As we have metioned,the non-local TV regularized softmax function can be described by Eq.(\ref{2.10}) and Eq.(\ref{2.11}).Solving this problem, the Min-max alternate algorithm is been required.

%\subsubsection{Min-Lagrange multiplier}
For \textbf{the $\mathscr{A}$-subproblem}, we employ the technique of Lagrange multipliers. The Lagrangian function is defined as:
\begin{equation}
\begin{gathered}
 L=\sum^{K}_{k=1}\{-\langle \mathscr{A}_{k}, O_{k}\rangle+\epsilon\langle\mathscr{A}_{k}, \log \mathscr{A}_{k}\rangle+\lambda\langle \mathscr{A}_{k}, \operatorname{div}_{\textbf{S}}\eta_{k}\rangle\}\\
 +\sum^{N}_{i=1}v_{i}\left(\sum^{K}_{k=1}\mathscr{A}_{ik}-1\right),   
\end{gathered}
\end{equation}
where $v_i$ are the Lagrange multipliers. Using the first order optimization conditions, we obtain:

\begin{equation}
\frac{\partial L}{\partial \mathscr{A}_{ik}}=-O_{ik}+\epsilon(\log \mathscr{A}_{ik}+1)+\lambda\operatorname{div}_{\textbf{S}}\eta_{ik}+v_{i}=0.
\end{equation}
Rearranging, we find:
\begin{equation}
\log \mathscr{A}_{ik}=\frac{O_{ik}-\epsilon-\lambda\operatorname{div}_{\textbf{S}}\eta_{ik}-v_{i}}{\epsilon}.
\end{equation}
Thus, we can express $\mathscr{A}_{ik}$ in the following form:
\begin{equation}
\mathscr{A}_{ik}=\exp\left(\frac{O_{ik}-\lambda\operatorname{div}_{\textbf{S}}\eta_{ik}}{\epsilon}\right)\exp\left(\frac{-\epsilon-v_{i}}{\epsilon}\right).
\label{A4}
\end{equation}
Considering the constraint
$
\sum^{K}_{k=1}\mathscr{A}_{ik}=1,
$
we have:
\begin{equation}
\exp\left({\frac{-\epsilon-v_{i}}{\epsilon}}\right)\sum^{K}_{k=1}\exp\left({\frac{O_{ik}-\lambda\operatorname{div}_{\textbf{S}}\eta_{ik}}{\epsilon}}\right)=1.
\end{equation}
Solving for $\exp({\frac{-\epsilon-v_{i}}{\epsilon}})$, we obtain:
\begin{equation}
\exp\left({\frac{-\epsilon-v_{i}}{\epsilon}}\right)=\frac{1}{\sum^{K}_{k=1}\exp\left({\frac{O_{ik}-\lambda\operatorname{div}_{\textbf{S}}\eta_{ik}}{\epsilon}}\right)}
\label{A6}.
\end{equation}
Substituting (\ref{A6}) into (\ref{A4}), we get:
\begin{equation}
\mathscr{A}_{ik}=\frac{\exp({\frac{O_{ik}-\lambda\operatorname{div}_{\textbf{S}}\eta_{ik}}{\epsilon}})}{\sum^{K}_{\hat{k}=1}\exp({\frac{O_{i\hat{k}}-\lambda\operatorname{div}_{\textbf{S}}\eta_{i\hat{k}}}{\epsilon}})}.
\end{equation}
Thus, we can express $\mathscr{A}_{k}$ as:
\begin{equation}
\mathscr{A}_{k}=\text{softmax}\left(\frac{O_{k}-\lambda\operatorname{div}_{\textbf{S}}\eta_{k}}{\epsilon}\right).
\end{equation}

%\subsubsection{Max-gradient descent}

For \textbf{the $\eta$-subproblem}, it is equivalent to solving the following minimization problem:

\begin{equation}
\min_{\|\eta_{k}\|_{\infty}\leq 1}\sum^{K}_{k=1}\langle\bigtriangledown_{\textbf{S}}\mathscr{A}_{k},\eta_{k}\rangle.
\end{equation}
We solve this using gradient descent and a projection operator:
 
\begin{equation}
\begin{aligned}
\eta^{t}_{k}=\prod\nolimits_{\|\eta_{k}\|_{\infty}\leq 1}(\eta^{t-1}_{k}-\tau \bigtriangledown \mathscr{A}^{t-1}_{k}).
\end{aligned}
\end{equation}

\end{document}